\documentclass[11pt,authoryear,review]{elsarticle}%,draft
\usepackage[final,commandnameprefix=ifneeded]{changes}
\definechangesauthor[name={Binbin},color=red]{R2}
\usepackage[bookmarks=true, colorlinks=true]{hyperref}
\usepackage{url}
\usepackage{algorithm}
\usepackage{algorithmicx}
\usepackage{multirow}
\usepackage{multicol}
\usepackage{amsfonts}
\usepackage{booktabs}
\usepackage{xcolor}
\usepackage{siunitx} 
\usepackage{float}
\usepackage{dsfont}
\usepackage{url}
\usepackage{ulem}
\usepackage{comment}
\usepackage[figuresright]{rotating}
\usepackage{amssymb,amsmath}
\usepackage{subfigure}
\usepackage{lineno}
\usepackage{tabularx}
\usepackage{longtable}
\usepackage{xltabular}
\usepackage{caption}
\usepackage{array}
\newcolumntype{C}[1]{>{\centering\arraybackslash}p{#1}}

\newcommand{\ks}[1]{\textcolor{orange}{\bf Konrad: #1}}

\journal{ISPRS Journal of Photogrammetry \& Remote Sensing}

\begin{document}
	
\begin{frontmatter}

\title{A Review of Panoptic Segmentation for Mobile Mapping Point Clouds}


\author{Binbin Xiang\corref{CorAuthor}}
\cortext[CorAuthor]{Corresponding author.}
\ead{binbin.xiang@geod.baug.ethz.ch}
\author{Yuanwen Yue}
\author{Torben Peters}
\author{Konrad Schindler}

\address{Photogrammetry and Remote Sensing, ETH Z{\"u}rich, 8093 Z{\"u}rich, Switzerland\\}
\begin{abstract}
3D point cloud panoptic segmentation is the combined task to (i) assign each point to a semantic class and (ii) separate the points in each class into object instances. Recently there has been an increased interest in such comprehensive 3D scene understanding, building on the rapid advances of semantic segmentation due to the advent of deep 3D neural networks. Yet, to date there is very little work about panoptic segmentation of outdoor mobile-mapping data, and no systematic comparisons. The present paper tries to close that gap. It reviews the building blocks needed to assemble a panoptic segmentation pipeline and the related literature. Moreover, a modular pipeline is set up to perform comprehensive, systematic experiments to assess the state of panoptic segmentation in the context of street mapping. As a byproduct, we also provide the first public dataset for that task, by extending the NPM3D dataset to include instance labels.\added[id=R2]{ That dataset and our source code are publicly available.\protect\footnote{https://github.com/bxiang233/PanopticSegForMobileMappingPointClouds} }\added[id=R2]{ We discuss which adaptations are need to adapt current panoptic segmentation methods to outdoor scenes and large objects.
Our study finds that for mobile mapping data, KPConv performs best but is slower, while PointNet++ is fastest but performs significantly worse. Sparse CNNs are in between. Regardless of the backbone, Instance segmentation by clustering embedding features is better than using shifted coordinates.}

%The main discoveries of our study reveal that among the deep learning backbones designed for point cloud processing, KPConv demonstrates superior performance for mobile mapping data, albeit with a slower processing speed. PointNet++ exhibits the highest speed but achieves the least impressive scores. Sparse CNN strikes a favorable balance in between. Regardless of the backbone used, instance segmentation by clustering embedding features outperforms clustering with shifted coordinates for our mobile mapping dataset. Moreover, certain modifications are required to effectively handle large outdoor objects like trees or buildings and to manage the processing and integration of extensive regions. These adjustments and their demonstration will be presented in this review.}
\end{abstract}


\end{frontmatter}

%\tableofcontents
%\newpage
%\linenumbers
%-------------------------------------------------------------------------
\section{Introduction}
\label{Sec:Introduction}
%-------------------------------------------------------------------------

Semantic segmentation and instance segmentation are two core tasks of scene understanding. Given densely sampled observations (e.g., pixels of an image, points of a point cloud), the goal of semantic segmentation is to assign a semantic category label to each observation. Instance segmentation aims to separate individual object instances, which only makes sense for categories that form clearly delineated objects. The joint task, to generate a complete, coherent scene interpretation in terms of semantic categories \emph{and} individual objects, has been termed \emph{panoptic segmentation}~\citep{Kirillov2019Panoptic}. In urban mobile mapping, panoptic segmentation has the natural application to structure raw 3D point clouds into semantically meaningful objects and surfaces, which are useful for higher-level tasks such as building inventories of street furniture, or enabling outdoor mobile robots. The aim of this paper is to establish a state of the art for the panoptic segmentation of mobile mapping point clouds, and to review the most promising algorithmic building blocks for that purpose.

\begin{figure}[htbp]
	\centering
	\includegraphics[width=\linewidth]{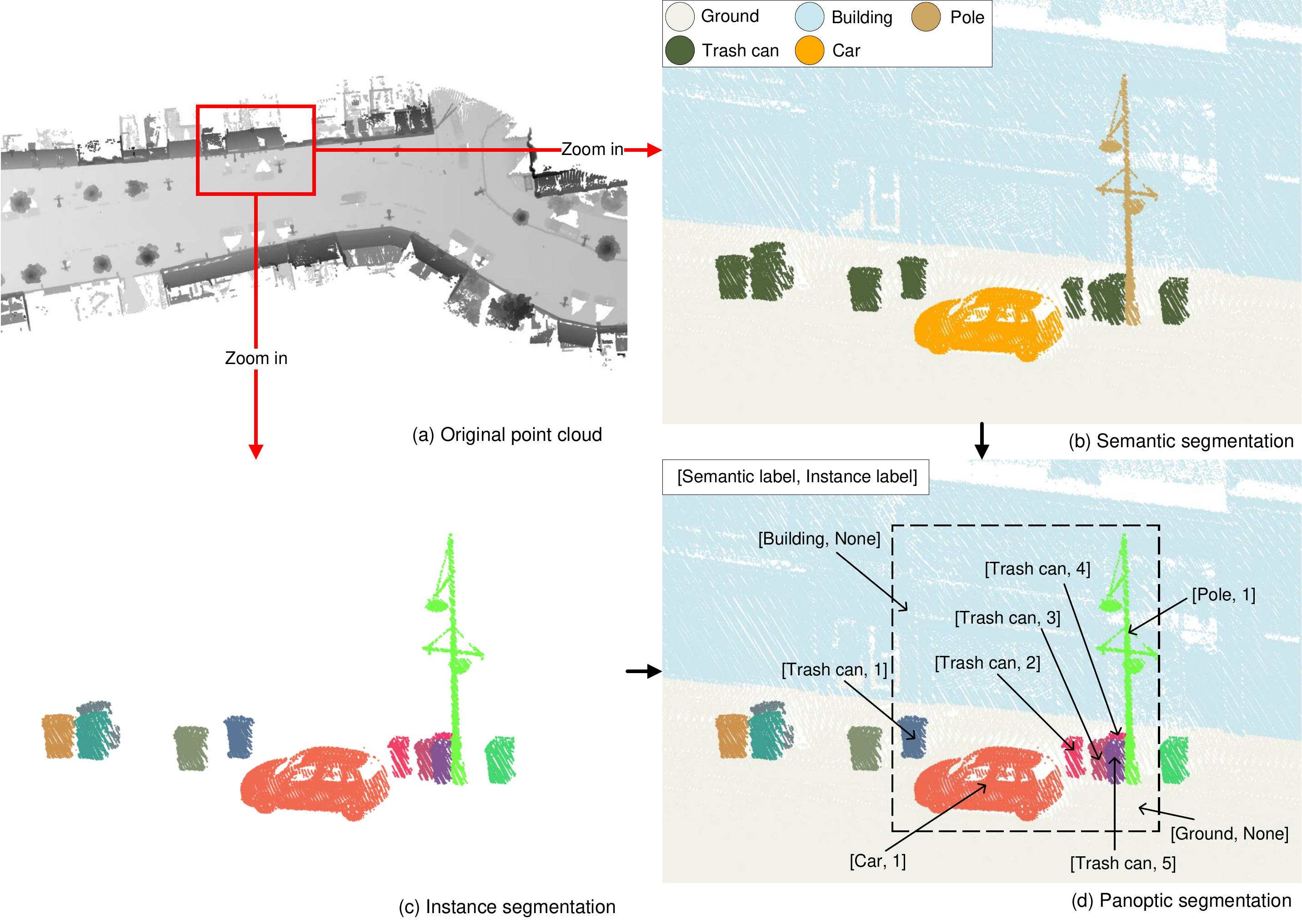}\\
	\caption{An input point cloud (a), its semantic segmentation (b, per-point semantic labels), instance segmentation (c, per-point instance labels, each label is assigned a random color), and panoptic segmentation (d, per-point semantic labels and, where applicable, instance labels). In the example, the categories ``ground'' and ``building'' are considered ``stuff'' and not separated into instances.}
	\label{Fig:PanopticSegExample}
\end{figure}

In panoptic segmentation, the compact, countable objects (such as pedestrians and cars) are often called ``things'', whereas the amorphous and uncountable regions (such as the road surface) are called ``stuff''. For ``stuff'' categories panoptic segmentation is the same as semantic segmentation, as they do not form instances: each point is assigned a semantically defined category label. Whereas for ``things'' the model must assign a semantic label and, additionally, a unique instance label that distinguishes it from other instances, but does not have a semantic meaning -- instance IDs are arbitrary and all perturbations of the labels are equally valid.  Figure~\ref{Fig:PanopticSegExample} illustrates the difference between semantic, instance, and panoptic segmentation.

\begin{figure}[!tb]
	\centering
	\includegraphics[width=\linewidth]{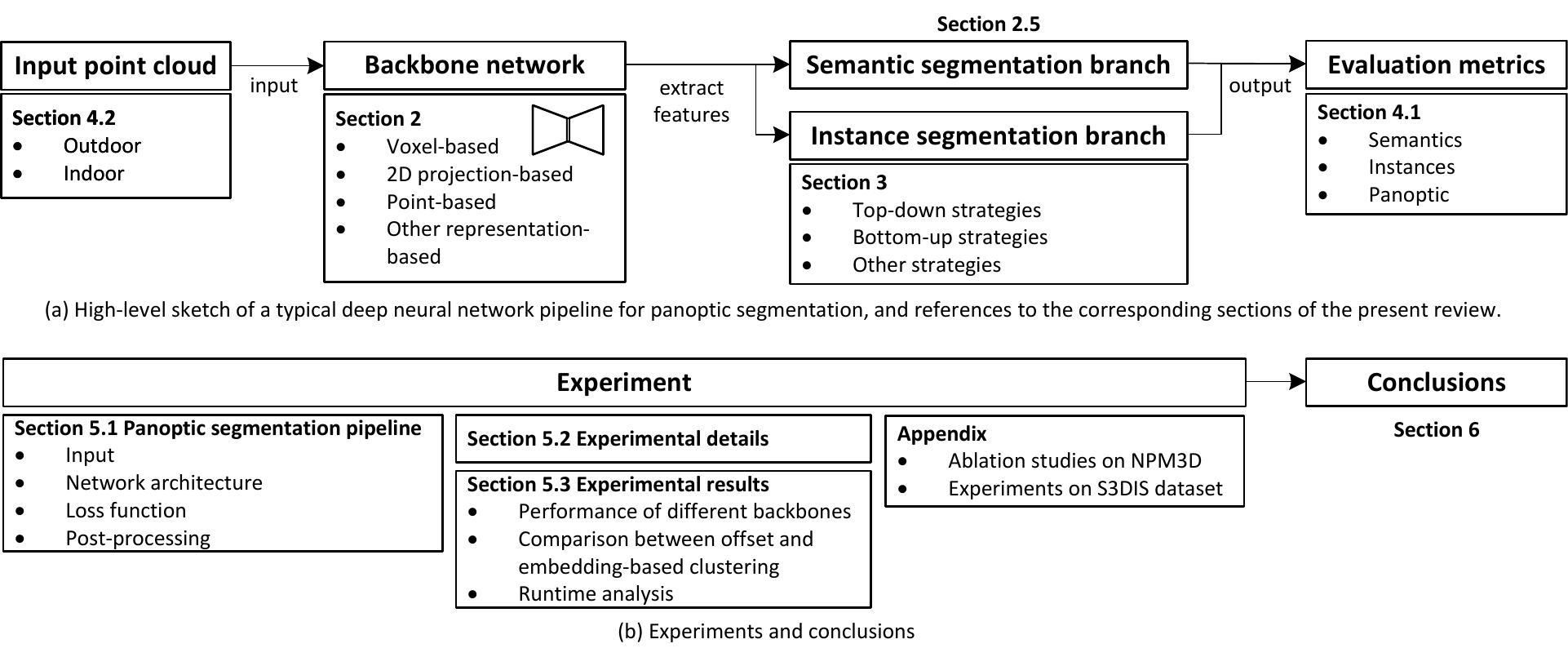}\\
	\caption{Complete overview of the paper.}
	\label{Fig:CompleteOverviewOfThePaper}
\end{figure}

As shown in Figure~\ref{Fig:CompleteOverviewOfThePaper}, panoptic segmentation is nowadays usually formulated as a combined classification and clustering task and solved with deep neural networks. The general architecture is to first extract a per-point feature encoding with a \emph{backbone network}, then feed the encoding into two parallel branches (heads), one that predicts semantic class probabilities and another one that further transforms the points such that points on the same instance form compact clusters\added[id=R2]{, achieving both task simultaneously based on shared features.} Historically, instance segmentation has been developed in 2D image analysis, as an add-on to semantic segmentation. Consequently, the focus has been on the design of the instance branch, without questioning the backbone. No systematic studies exist that investigate how different 3D backbones support point cloud instance segmentation. Moreover, panoptic segmentation of outdoor scenes has almost exclusively been studied for sparse, sequential scan data from autonomous driving scenarios like SemanticKITTI~\citep{behley2019iccv, behley2021ijrr} or nuScenes~\citep{nuscenes2019,fong2021panoptic}\added[id=R2]{, often in a real-time sceneario.} It appears that panoptic segmentation of dense, mapping-grade point clouds has not been investigated much. 

\begin{comment}
\begin{figure}[htbp]
	\centering
	\includegraphics[width=\linewidth]{figures/Topics surveyed in this review}\\
	\caption{Topics discussed in this review.}
	\label{Fig:TopicsSurveyedInThisReview}
\end{figure}
\end{comment}

There are a number of review papers about deep learning for 3D point clouds.
Some review papers attempt a comprehensive overview of different point cloud analysis tasks\replaced[id=R2]{~\citep{Ioannidou2017DeepLA,Liu2019Deep,Bello2020review,Deep2020Li,Lu2020Deep,Guo2020Deep}}{\mbox{~\citep{Ioannidou2017DeepLA,Guo2020Deep,Lu2020Deep,Deep2020Li,Liu2019Deep,Bello2020review}}}, predominantly scene-level classification, segmentation, and object detection. Others concentrate on a single analysis task
\replaced[id=R2]{~\citep{Griffiths2019Areview,Wu2021Deep,He2021Deep,Zamanakos2021Comprehensive,zhao2021technical,Burume2021Deep,Jhaldiyal2022Semantic,Diab2022Deep,Alaba2022ASurvey}}{\mbox{~\citep{Zamanakos2021Comprehensive,Griffiths2019Areview,Alaba2022ASurvey,zhao2021technical,Wu2021Deep,He2021Deep,Diab2022Deep,Burume2021Deep,Jhaldiyal2022Semantic}}}. For example, several papers review 3D object detection methods for autonomous driving\replaced[id=R2]{~\citep{Wu2021Deep,Zamanakos2021Comprehensive,Alaba2022ASurvey}}{\mbox{~\citep{Zamanakos2021Comprehensive,Alaba2022ASurvey,Wu2021Deep}}}, and~\cite{Burume2021Deep} focus on instance segmentation in that scenario.~\cite{zhao2021technical} did a survey to explore the effects of different traditional point cloud clustering algorithms for LiDAR panoptic segmentation task. To date, there has not been a review that specifically addresses the panoptic segmentation task for dense, mapping-grade 3D LiDAR point clouds. The present work aims to fill that gap. Besides systematically introducing the task and discussing important representative works, it also provides a unified experimental test bed and comparative results for reference.

Following the literature about deep learning for point clouds\replaced[id=R2]{~\citep{Griffiths2019Areview,Deep2020Li,Wu2021Deep,He2021Deep,Zamanakos2021Comprehensive,Diab2022Deep,Alaba2022ASurvey}}{\mbox{~\citep{Zamanakos2021Comprehensive,Griffiths2019Areview,Alaba2022ASurvey,Wu2021Deep,He2021Deep,Diab2022Deep,Deep2020Li}}}, 3D backbone networks are classified into four classes: voxel-based, point-based, 2D projection-based, and other presentations, which will be discussed in Section~\ref{Sec:3DBackboneNetworksImprovements}. For the further investigation of dense outdoor point cloud panoptic segmentation, the paper focuses on three representative backbone networks: PointNet++~\citep{Qi2017PointNetDH}, Sparse CNN~\citep{choy20194d}, and KPConv~\citep{Thomas2019KPConvFA}. The choice is motivated by the following considerations. First, PointNet++ and Sparse CNN are currently the most widely used point cloud backbones (see Table~\ref{table:summarizesmethods}). Second, KPConv and Sparse CNN are particularly successful backbones for segmentation tasks, with KPConv achieving state-of-the-art performance for semantic segmentation on the NPM3D dataset~\citep{roynard2017parislille3d}, while Sparse CNN is the most widely used backbone for state-of-the-art instance segmentation of indoor point clouds~\citep{jiang2020pointgroup,chen2021hierarchical,Vu2022SoftGroup}. The three types of backbones are described in detail in Section~\ref{Sec:3DBackboneNetworksImprovements}.

Section~\ref{sec:StrategiesForInstanceSegmentationInPanopticSegmentation} of the paper distinguishes top-down and bottom-up strategies, like other  articles about 3D point cloud instance segmentation~\citep{He2021Deep,Burume2021Deep}. Additionally, the present paper introduces and summarizes the two most commonly used strategies for bottom-up methods (Table~\ref{table:summarizesmethods} and Section~\ref{subsubSec:LossFunctions}), and compares their performance in the context of outdoor panoptic segmentation (Section~\ref{subsec:2featuresForInstanceClustering}).

So far, there is no outdoor mobile mapping dataset tailored to panoptic segmentation, as existing datasets lack instance labels. Existing outdoor datasets of LiDAR point clouds with instance labels are dominated by the autonomous driving setting. I.e., the points are comparatively sparse, moreover instance labels are typically limited to pedestrians and vehicles (see Section~\ref{subSec:Outdoordataset}), whereas for mapping a wider variety of object classes should be instantiated, such as trees, poles, street lamps and so on. For the present work, this meant on the one hand that a new dataset had to be generated, which we do by adding instance labels to an existing benchmark, see Section~\ref{subSec:Outdoordataset}). On the other hand, the lack of a common dataset also meant that for the review it was not possible to collect the results of different methods from the literature. Instead, a common code base for different methods was implemented and applied on the new dataset to enable informative and fair comparisons.

%\newpage
In summary, the \emph{main contributions} of the present paper are:
\begin{itemize}
\item A detailed literature review about 3D point cloud panoptic segmentation, grouped into four sub-topics: datasets, 3D backbone networks for semantic segmentation, strategies for instance segmentation, and evaluation metrics. See also Figure~\ref{Fig:CompleteOverviewOfThePaper}.
\item An experimental evaluation and comparison in the outdoor mobile mapping setting. The evaluation concentrates on the bottom-up segmentation and grouping strategy (see Section~\ref{sec:StrategiesForInstanceSegmentationInPanopticSegmentation}), which so far proved to be the most successful approach and dominates the leaderboards for indoor datasets. To that end, a modular panoptic segmentation pipeline (Section~\ref{subSec:ProposedPipeline}) was created that allows one to combine different feature extraction and segmentation strategies. Extensive experiments are run on the NPM3D dataset~\citep{roynard2017parislille3d} with three different, representative backbones and two representative instance clustering approaches (Section~\ref{Sec:ExperimentalResults}).
\item The instance-annotated dataset and all source code will be published, so as to expedite further research and promote the development of 3D panoptic segmentation. So far, there do not seem to be any other works that systematically investigate modern deep learning methods for panoptic segmentation of mobile mapping point clouds. 
\end{itemize}

\begin{comment}
\begin{figure}[htbp]
	\centering
	\includegraphics[width=\linewidth]{figures/Topics surveyed in this review}\\
	\caption{Topics discussed in this review.}
	\label{Fig:TopicsSurveyedInThisReview}
\end{figure}
\end{comment}

%-------------------------------------------------------------------------
\section{Backbone networks for 3D point cloud analysis}
\label{Sec:3DBackboneNetworksImprovements}
%-------------------------------------------------------------------------
Traditionally, 3D point cloud analysis has relied on manually designed features that capture the local point distribution (and thus, the surface shape) around a point~\citep{niemeyer2012conditional, Li2016ATA, Zhu2017RobustPC}. This works well for categories with sufficiently obvious, distinctive features, but much time and effort are required to design features, select appropriate neighborhoods, etc. Such approaches no longer reach state-of-the-art performance, mostly because it is not clear how to manually design a good representation of contextual relations over larger contexts. Furthermore, the handcrafted features tuned for a point cloud with specific sensing characteristics typically fail to handle other types of scenes or other sensor settings, meaning that the tedious manual feature engineering must be repeated.

At present, deep learning is the standard technology to automatically extract discriminative, robust feature representations from raw data. Remarkable results have been achieved for 2D image classification, semantic segmentation, object detection, and more~\citep{lecun2015deep}. Motivated by those results and a growing number of public 3D point cloud datasets, researchers have developed powerful deep learning algorithms to also extract features for 3D point cloud analysis~\citep{Ioannidou2017DeepLA}. These existing 3D ``backbone'' networks can be grouped into three main schools, namely voxel-based networks, 2D projection-based networks, and point-based networks. Representative examples are summarized in Table~\ref{table:compare_backbones}.

%-------------------------------------------------------------------------
\subsection{Voxel-based networks}
\label{subSec:VolumetricCNN}
%-------------------------------------------------------------------------

\begin{figure}[htbp]
	\centering
	\includegraphics[width=\linewidth]{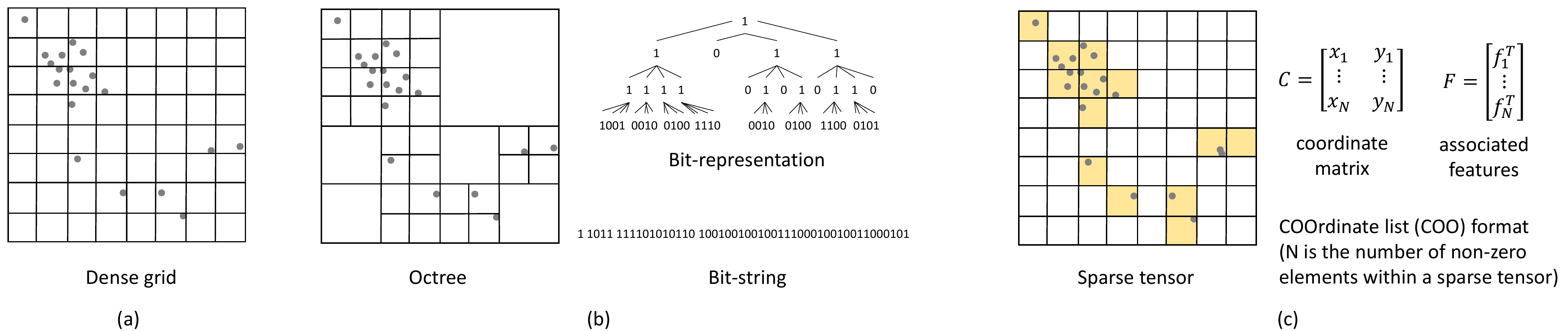}\\
	\caption{Voxel-based representations. Three voxel-based data structures for point clouds are compared: dense grid, octree, and sparse tensor. For clarity, they are illustrated in 2D space. (a) Dense grids correspond to explicitly storing all voxels as a dense tensor (in 2D, a matrix). For the octree, like the one shown in (b), space is hierarchically subdivided, splitting only voxels that contain points. The resulting tree can be efficiently represented by a bit-string. The sparse tensor uses the COO format as shown in (c), with one matrix that stores the coordinates of occupied voxels and another one storing the feature values only for those voxels (yellow areas).}
	\label{Fig:voxel-basedrepresentations}
\end{figure}

\begin{figure}[htbp]
	\centering
	\includegraphics[width=\linewidth]{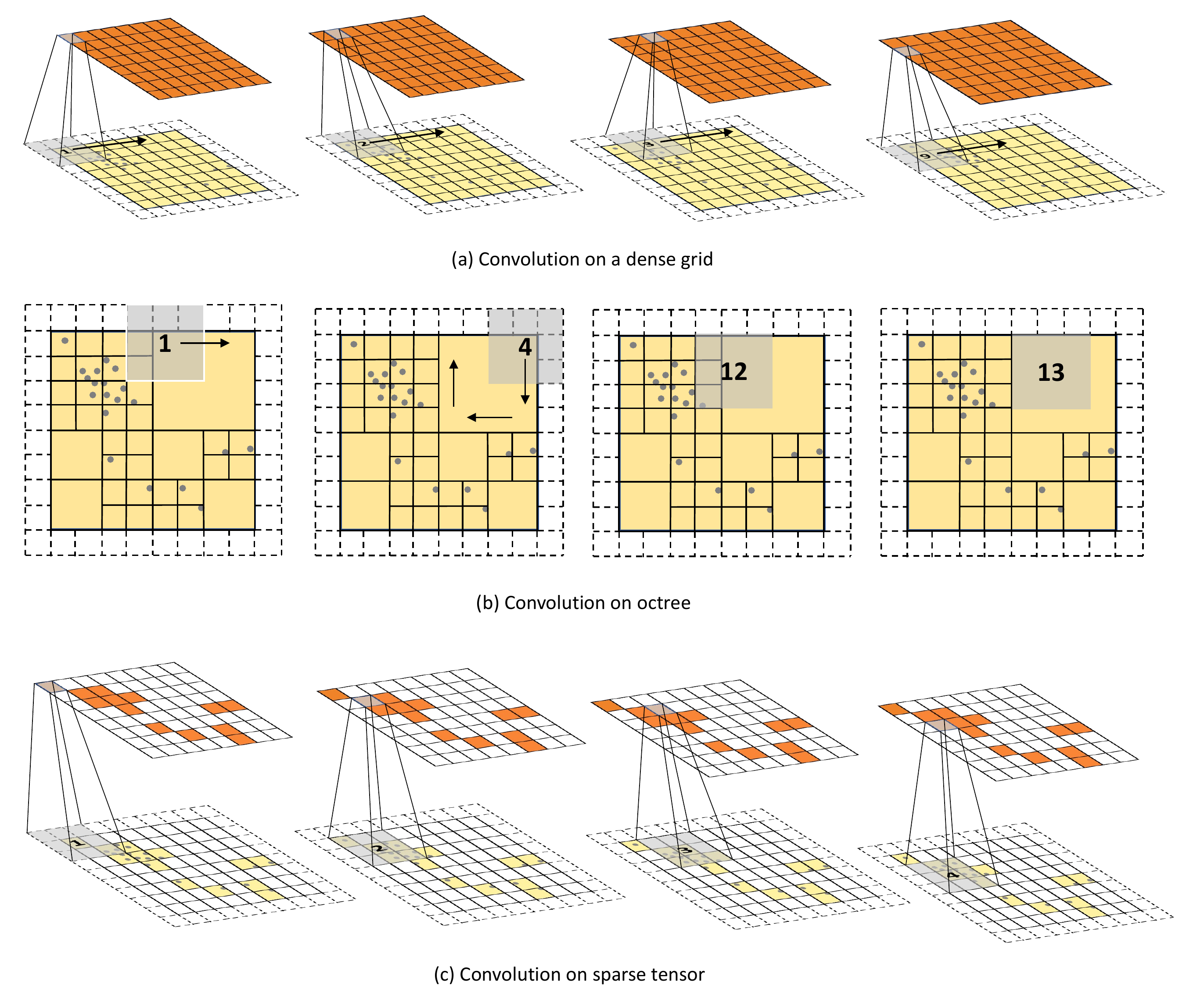}\\
	\caption{\added[id=R2]{Illustration of voxel-based convolutions in different formats. For clarity, they are shown in 2D space. Arrows indicate the direction in which the convolution kernel is moved, numbers denote the kernels' sequence order. (a) On a full, dense voxel grid standard discrete convolutions are applied in 3D. (b) The octree evaluates the kernel only along the borders of large voxels, where the support changes due to adjacent voxels. (c) On sparse tensors convolutions are evaluated only at non-empty locations.}}
	\label{Fig:voxel-basedconvolutions}
\end{figure}

The points in a point cloud are generally unstructured and have irregular density. These characteristics are quite different from images, where relationships between pixels can be captured with regular, discrete convolution kernels. 
The simplest and most direct strategy to apply deep learning to 3D point clouds is to voxelise the underlying point data and use 3D Convolutional Neural Networks (CNNs)~\citep{Maturana2015VoxNetA3} (Figure~\ref{Fig:voxel-basedrepresentations}(a) and Figure~\ref{Fig:voxel-basedconvolutions}(a)). Perhaps the first attempt to use CNNs for semantic segmentation of outdoor point clouds was~\cite{Huang2016Point}. However, when moving to a voxel representation the computational cost and, more importantly, the memory demand increases cubically with the resolution, thus limiting the practical usability of 3D CNNs~\citep{Riegler2017OctNetLD}. To bring down the memory requirements, hierarchical and/or sparse voxel representations have been developed. Hierarchical volumetric models~\citep{Riegler2017OctNetLD,Wang2017OCNNOC} use data structures like octrees to focus memory allocation and computation on the relevant regions where there are points (Figure~\ref{Fig:voxel-basedrepresentations}(b) and Figure~\ref{Fig:voxel-basedconvolutions}(b)). The same principle, to omit the vast amount of empty 3D space, is also the basis of methods that employ sparse tensors~\citep{Graham20183DSemantic,choy20194d,tang2020searching} and associated functions that only perform calculations for non-empty voxels (Figure~\ref{Fig:voxel-basedrepresentations}(c) and Figure~\ref{Fig:voxel-basedconvolutions}(c)). The Minkowski Engine~\citep{choy20194d} and the SparseConvNet library~\citep{Graham20183DSemantic} are two popular software frameworks that support all standard neural network layers like convolution, pooling and upsampling, for sparse tensors. Figures~\ref{Fig:voxel-basedrepresentations} and~\ref{Fig:voxel-basedconvolutions} illustrate how dense voxel grids, octrees and sparse tensors, respectively, store point clouds and compute  convolutions.

\begin{comment}
\begin{figure}
  \subfigure[3D CNN based semantic segmentation pipeline for outdoor large-scale data~\citep{Huang2016Point}]{
  \begin{minipage}[t]{1\linewidth}
    \centering
    \includegraphics[scale=0.7]{figures/reviews/voxel-based paper.png}
    \vspace{-1.2cm}
    \label{Fig:voxel-basedpaper}
  \end{minipage}%
  }
  \hspace{.15in}
  \quad
  \vspace{1.5cm}
  \subfigure[Take a voxelized bed as an example, the second row and third row show difference between dense 3D CNN and OctNet representations~\citep{Riegler2017OctNetLD}.]{
  \begin{minipage}[t]{0.5\linewidth}
    \centering
    \includegraphics[scale=0.5]{figures/reviews/Octnet.png}
    \vspace{-1cm}
    \label{Fig:Octnet}
  \end{minipage}%
  }
  \hspace{.15in}
  \vspace{1.5cm}
  \subfigure[An illustration of 2D convolution on a dense tensor and a sparse tensor~\citep{choy20194d}.]{
  \begin{minipage}[t]{0.5\linewidth}
    \centering
    \includegraphics[scale=0.3]{figures/reviews/dense vs sparse tensor}
    \vspace{-1cm}
    \label{Fig:denseVSSparseTensor}
  \end{minipage}
  }
  \vspace{-3.2cm}
  \caption{Voxel-based backbones.\ks{I don't think we need this figure, also if it is from a published paper there might be copyright issues with reprinting it. Perhaps just leave it out?}}
  \label{fig:voxel-basedbackbones}
\end{figure}
\end{comment}

%-------------------------------------------------------------------------
\subsection{2D projection-based networks}
\label{subSec:2DProjection-basedNetworks}
%-------------------------------------------------------------------------

To avoid the complications of a 3D voxel space, another strategy are 2D projection-based methods. Their basic idea is to convert the original 3D point cloud into a regular 2D raster, which can then be treated with standard 2D CNNs. Variants include single- or multi-view perspective projections~\citep{Lawin2017DeepP3,Kalogerakis20173DSS}, parallel~\citep{Tatarchenko2018TangentCF} as well as spherical projections~\citep{wu2018squeezeseg,wu2019squeezesegv2}. This type of approach tends to work well when the underlying assumptions are met, e.g., well-behaved surfaces with an unambiguous normal/tangent~\citep{Tatarchenko2018TangentCF} or individual scans (respectively, range images) that, in actual fact, induce a 2D parametrisation.

%-------------------------------------------------------------------------
\subsection{Point-based networks}
\label{subSec:PointBasedNetworks}
%-------------------------------------------------------------------------

\begin{figure}[htbp]
	\centering
	\includegraphics[width=\linewidth]{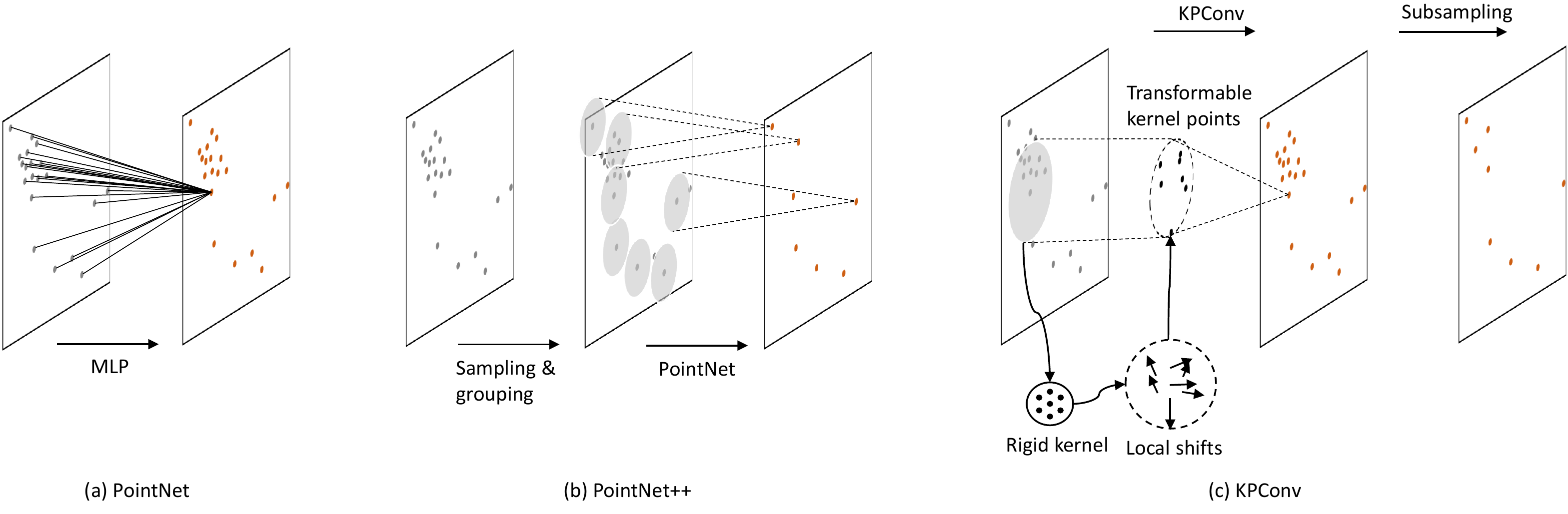}\\
	\caption{Illustration of three point-based backbones. For clarity, the 2D versions are depicted. \added[id=R2]{(1) In PointNet, each input point is processed separately through a shared MLP. All points are aggregated through global pooling. (2) PointNet++ improves upon PointNet with sampling and grouping operations to create local point sets and gradually aggregate their features. As a result, PointNet++ generates subsampled point clouds with enhanced representations, enabling efficient processing of large-scale data while preserving important details and global context. (3) KPConv extends traditional convolutions by defining the kernel weights at arbitrary support points, which additionally can be shifted locally to adapt to the local point distribution. The computation involves aggregating information from neighboring points within a fixed receptive field, enabling KPConv to extract local features. Additionally, KPConv incorporates subsampling methods to obtain multi-scale features.}}
	\label{Fig:point-basedbackbones}
\end{figure}

To side-step the representation change altogether, point-based networks directly work on irregular point clouds\replaced[id=R2]{~\citep{Qi2017PointNetDL,Qi2017PointNetDH,hua2018pointwise,wang2019graph,Wang2019DynamicGC,Thomas2019KPConvFA}}{\mbox{~\citep{Qi2017PointNetDL,Qi2017PointNetDH,hua2018pointwise,Wang2019DynamicGC,wang2019graph,Thomas2019KPConvFA}}}. The earliest such method was PointNet~\citep{Qi2017PointNetDL}, where a per-point multiple-layer perceptron (MLP) is combined with max-pooling to achieve a global feature invariant to permutations of the points (Figure~\ref{Fig:point-basedbackbones}(a)). PointNet++~\citep{Qi2017PointNetDH} builds upon PointNet and adds a step-wise hierarchical aggregation to retain more of the spatial layout (Figure~\ref{Fig:point-basedbackbones}(b)). A different method, more in the spirit of CNNs, is the kernel point convolution~\citep[KPConv,][]{Thomas2019KPConvFA}, which approximates the 3D convolution in continuous space by interpolation and also hierarchically aggregates multi-scale features to capture both local and global information (Figure~\ref{Fig:point-basedbackbones}(c)). It has shown excellent performance on outdoor data, like NPM3D~\citep{roynard2017parislille3d} and Semantic3D~\citep{hackel2017isprs}.

%-------------------------------------------------------------------------
\subsection{Other representations}
\label{subSec:OtherRepresentationBasedNetworks}
%-------------------------------------------------------------------------

Besides the three main groups described so far, there have been a few other attempts to process 3D data with reasonable memory footprint. These include permutohedral lattices~\citep{su2018splatnet,rosu2019latticenet}, offline grouping into irregular patches, combined with graph convolutional networks~\citep{landrieu2018large}, and hybrid models~\citep{dai20183dmv,jaritz2019multi}.

\subsection{Semantic segmentation of point clouds}
Semantic segmentation assigns each point to a semantic class, without distinguishing different instances within the class. This is accomplished by adding a segmentation ``head'' to the backbone, which outputs a class distribution for each individual point. Overall, this results in an encoder-decoder structure: the backbone encodes the input into a latent representation optimised for classification, and the classification head decodes that representation into class scores. For voxel-based, projection-based and sparse tensor-based methods, the decoder consists of a sequence of transposed convolutions. For PointNet-style methods, the per-point features and the latent representation over a larger spatial context are simply concatenated and decoded with an MLP.

\newpage
{
\tiny
\begin{xltabular}{\textwidth}{m{0.2cm}|p{3cm}p{3.5cm}p{3.6cm}p{3.6cm}}
\captionsetup{justification=centering}
\caption{Summary of different types of representative backbone networks for 3D point clouds.} \label{table:compare_backbones} 
  \\ \hline
  &
  \multicolumn{1}{c}{Representatives} &
  \multicolumn{1}{c}{Methodology}&
  \multicolumn{1}{c}{Advantages} &
  \multicolumn{1}{c}{Disadvantages}  \\ \hline
  \endfirsthead

  \multicolumn{5}{c}%
  {\tablename\ \thetable{} -- continued from previous page} \\ \hline
  &
  \multicolumn{1}{c}{Representatives} &
  \multicolumn{1}{c}{Methodology}&
  \multicolumn{1}{c}{Advantages} &
  \multicolumn{1}{c}{Disadvantages} \\ \hline 
  \endhead

  %\hline 
  \multicolumn{5}{r}{{Continued on next page}} \\ %\hline
  \endfoot
  \hline
  \endlastfoot
  
  \multirow{3}{*}[-65ex]{\rotatebox[origin=c]{90}{Voxel-based}}&
  3D CNN~\citep{Maturana2015VoxNetA3,Huang2016Point} &
  Encode 3D shapes by occupancy voxels. &
  - \underline{Simplicity:} Direct application of 3D CNNs to 3D point clouds in voxel grids.\newline
  - \underline{Searching} for neighboring points takes only $\mathcal{O}(1)$ operations.
  &
  - \underline{High computational cost} due to large number of voxels. Requirements increase cubically by $\mathcal{O}(n^3)$ with resolution.\newline
  - \underline{Low efficiency} due to high sparsity of the voxel grid.  
  \newline
  - \underline{Quantization errors} due to voxelization of the point cloud.   \\ \cline{2-5} 
  % is actually unnecessary.
  %\newline - \underline{Hard parameters selection} due relationship between The grid size is difficult to adjust, which affects the scale of the input data and may destroy the spatial relationship between points. \\ \cline{2-5} 
  &
  - OctNet~\citep{Riegler2017OctNetLD}\newline- O-CNN~\citep{Wang2017OCNNOC} &
  Octree instead of grid representation. Hierarchically divide 3D space into a set of unbalanced octants based on point cloud density. &
  -~\underline{Much more efficient:} Storage and computational time reduces to $\mathcal{O}(n^2)$ compared to grid-based methods.\newline-~\underline{Improved performance} for high-resolution 3D data when compared to dense voxel-based (3D CNN) models. &
  -~\underline{Slower search time} compared to voxel grids with $\mathcal{O}(log(n))$ due to superimposed tree structure.  
  \newline-~\underline{Input size limitation}. Due to the restriction of the maximal depth of an octree (i.e., three) \citep{Xiang2019Novel}, hindering proper grid size for large-scale applications.\\ \cline{2-5} 
  %\multirow{1}{*}[-80ex]{\rotatebox[origin=c]{90}{\makecell{Voxel-based}}}
  &
  Sparse tensor CNN~\citep{Graham20183DSemantic,choy20194d} &
  %- Inspired by O-CNN and improve on the spatial sparsity of high-dimensional data.\newline- 
  An N-dimensional extension of a sparse matrix that only saves information on the non-empty region of the space. 
  &
  -~\underline{Computational efficiency:} Fast computation due to reduced computations on unoccupied elements in the voxel grid. 
  \newline-~\underline{Memory efficiency:} 3D sparse tensors enable the representation of large sparse arrays in a compact format, reducing memory requirements.

  %Is this really a benefit? As far as I understand higher accuracy boils down to less memory requriments, which allows more complex models. 
  
  %\newline-~\underline{Higher accuracy}. The ability to use higher resolution voxelization and deeper networks results in increased accuracy. Sparse representations can also improve accuracy by lowering the impact of irrelevant or noisy data. 

  %is this important ?
  %\newline-~\underline{Optimization of batching.} Assemble heterogeneous batches by directly collating the samples in the dimension of the point and keeping track of the number of points in each sample (packed batching). It eliminates the need to resample the batch elements, and in some cases can significantly reduce the memory requirements. 
  -~\underline{Highly optimized frameworks}. Minkowski Engine~\citep{choy20194d} and SparseConvNet, which support all standard CNN operations \citep{Graham20183DSemantic} 
  &\underline{Quantization errors}. All voxel-based methods lose some geometric properties of 3D objects, particularly the intrinsic characteristics of patterns and surfaces. \\ \hline
  
  \multirow{1}{*}[-4ex]{\rotatebox[origin=c]{90}{2D projection-based}}&
  - Multi views~\citep{Lawin2017DeepP3}\newline- Bird’s-eye view~\citep{Zhang2018Efficient}\newline- Spherical projection~\citep{wu2018squeezeseg,wu2019squeezesegv2} 
  &
  %Transform the points in 3D space into 2D regular images as the network input by projecting them in a certain way. 
  Projects 3D point clouds into images and applies 2D CNNs to them. 
  &
  -~\underline{2D network utilization:}. Fully compatible with well-established 2D CNNs and pretrained networks on image datasets. 
  \newline-~\underline{Efficiency:} Same requirements as 2D CNNs with $\mathcal{O}(n^2)$ time and space complexity.\added[id=R2]{ Suitable for real-time applications.}   
  &
  -~\underline{Occlusions and Distortions:} Many projection-related errors appear, including various types of occlusions and projective distortions. 
  \newline-~\underline{Hyperparameters:} Adds additional hyperparameters for camera and poses. 
  \\ \hline
  \multirow{1}{*}[-12ex]{\rotatebox[origin=c]{90}{Point-based}}&
  PointNet~\citep{Qi2017PointNetDL} &
 Learns to project 3D points together with point attributes independently into a common feature space; Features are then aggregated using pooling for further processing. &
 -~\underline{No preprocessing required:} Works directly on point clouds and is invariant to point ordering.
 \newline-~\underline{Efficiency:} Very low time and space complexity of $\mathcal{O}(n)$, which increases linear with the number of points.
  &
  \underline{Inaccurate:} PointNet cannot capture fine-grained patterns and local structure, especially in large scenes. Independent feature representation of entire point cloud is aggregated into a single vector, resulting in a loss of information. 

 \\ \cline{2-5} 
  \multirow{2}{*}[-38ex]{\rotatebox[origin=c]{90}{Point-based}}&
  PointNet++~\citep{Qi2017PointNetDH} &
  Hierarchical application of mini PointNets on local 3D neighborhoods. Every layer applies PointNets on local 3D regions, which are then aggregated, grouped and passed to the following layer.&
  \underline{Multi-scale features}. It extracts local features for points at different scales, thus improving the accuracy over PointNet. &
  -~\underline{Information loss}. Although it treats points independently at local scales in order to maintain permutation invariance, this independence ignores the geometric relationships between points and their neighbors, resulting in the absence of local features.
  %\newline- Local neighborhood points in different sampling layers are learned independently. The maxpooling operation for high-level feature extraction based on PointNet fails to preserve spatial information between local neighboring points.
  \newline-~\underline{Relative complexity}. It is more complex than PointNet and has more parameters. It is 3 times (or more) slower than PointNet. \\ \cline{2-5} 
  &
  KPConv~\citep{Thomas2019KPConvFA} &
  KPconv is a point convolution in 3D. It first builds a radius graph pyramid on the 3D input point cloud by a combination of local neighborhood search and regular grid subsampling. Then the graph is processed by applying spatial convolution on the local $k$ neighbors of each 3D point. It includes a version with a deformable kernels. 
  &
  -~\underline{No preprocessing required:} Works directly on point clouds.
  \newline-~\underline{Accuracy}. Outperforms traditional convolutional networks in 3D object classification and segmentation tasks.
  \newline-~\underline{Density Robustness:} It is more robust to varying densities because of a regular subsampling strategy. 
  &
  \underline{Computational complexity}. KPConv has a higher computational complexity due to the local neighborhood search.\\ %\hline
\end{xltabular}
}

%-------------------------------------------------------------------------
\section{Strategies for instance segmentation}\label{sec:StrategiesForInstanceSegmentationInPanopticSegmentation}
%-------------------------------------------------------------------------

\begin{figure}[!tb]
	\centering
	\includegraphics[width=\linewidth]{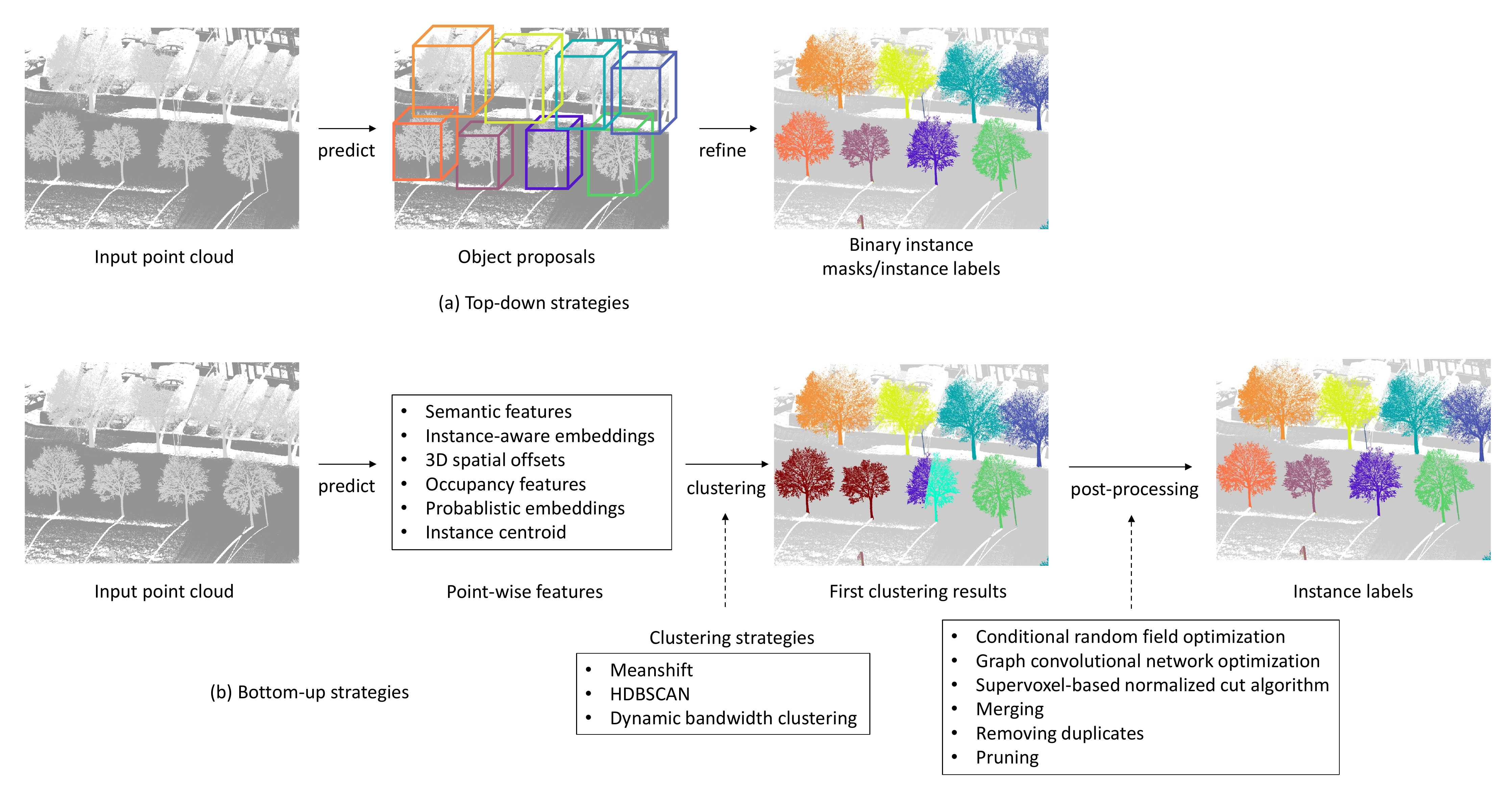}\\
	\caption{\added[id=R2]{Illustration of top-down (first row) and bottom-up (second row) strategies for instance segmentation.}}
	\label{Fig:instanceSegmentationStrategies}
\end{figure}

Once reliable backbones for 3D point cloud analysis were available, a natural next step was to also look into instance segmentation. This Section reviews the literature on that topic. At a conceptual level, existing strategies for instance segmentation can be grouped into two main types: top-down and bottom-up strategies. Figure~\ref{Fig:instanceSegmentationStrategies} illustrates the difference between the two strategies.

%-------------------------------------------------------------------------
\subsection{Top-down strategies}
\label{subSec:Top-downStrategies}
%-------------------------------------------------------------------------

Top-down methods\replaced[id=R2]{~\citep{Yi2019GSPNGS,Yang2019LearningOB,Zhang2020InstanceSO,liu2020learning}}{\mbox{~\citep{Yi2019GSPNGS,Yang2019LearningOB,liu2020learning,Zhang2020InstanceSO}}}, sometimes also termed proposal-based methods, resemble the Mask R-CNN~\citep{He2017MaskR} approach to object detection (Figure~\ref{Fig:instanceSegmentationStrategies}(a)). Instance segmentation is implemented as a sequence of two steps. First, localize a 3D bounding box of each instance, with a standard object detector that constructs candidate boxes and classifies them. Then predict per-point binary instance masks within each bounding box to obtain the final instance segmentation. Instead of directly regressing 3D bounding boxes,\replaced[id=R2]{~\cite{Yi2019GSPNGS}}{ Yi et al.\mbox{~\citep{Yi2019GSPNGS}}} design a generative model named Generated Shape Proposal Network (GSPN) to produce 3D proposals with a high probability of being compact objects. GSPN is integrated into an instance segmentation pipeline called Region-based PointNet (R-PointNet) that can flexibly refine proposals generated with GSPN into segmentation masks. Unlike R-PointNet, which needs two-stage training, 3D-BoNet~\citep{Yang2019LearningOB} is an end-to-end network where bounding box detection and instance masking can be trained simultaneously. It also avoids intensive post-processing (non-maximum suppression) to prune dense object proposals. For large-scale outdoor point clouds,~\cite{Zhang2020InstanceSO} proposed a simple top-down instance segmentation that encodes a 2D grid-level feature representation in a birds-eye view and predicts the planimetric object center and the height limits for each foreground grid location. Then it merges locations with similar object centers into instances.

Proposal-based methods are less prone to over-segmenting instances, thanks to the explicit object detector. The flip-side is that they cannot recover missed detections, and are challenged by overlapping detections. Typically their multi-stage inference and post-processing also lead to comparatively slow runtimes.

%-------------------------------------------------------------------------
\subsection{Bottom-up strategies}
\label{subSec:Bottom-upStrategies}
%-------------------------------------------------------------------------

\begin{comment}
\begin{figure}[!tb]
	\centering
	\includegraphics[width=\linewidth]{figures/reviews/Bottom-up strategies.pdf}\\
	\caption{High-level illustration of the bottom-up instance segmentation strategies.}
	\label{Fig:Bottom-upStrategies}
\end{figure}
\end{comment}

Bottom-up instance segmentation strategies (also called proposal-free or clustering-based methods, as shown in Figure~\ref{Fig:instanceSegmentationStrategies}(b)) tend to outperform top-down methods, and have thus become popular\replaced[id=R2]{~\citep{Wang2018SGPNSG,Liu2019MASCMA,Liang20193DGE,Wang2019AssociativelySI,Pham2019JSIS3DJS,Elich20193DBEVISBI,Lahoud20193DIS,han2020occuseg,zhang2020ssen,Wang2020Hierarchical}}{\mbox{~\citep{Wang2018SGPNSG,Wang2019AssociativelySI,Pham2019JSIS3DJS,Lahoud20193DIS,Liu2019MASCMA,Elich20193DBEVISBI,Liang20193DGE,zhang2020ssen,han2020occuseg}}}. Their basic principle is to map points to a discriminative representation space where points from the same instance have similar features, whereas points on different objects have dissimilar features. Instances are then retrieved by clustering in that space\added[id=R2]{~\citep{Comaniciu2002Mean,Campello2013Density,Neven2019Instance,hong2021lidar}, as illustrated in Figure~\ref{Fig:instanceSegmentationStrategies}(b)}. In the perhaps earliest attempt, SGPN~\citep{Wang2018SGPNSG} learns a similarity matrix that indicates pairwise similarities between points, then merges points with high similarity with a heuristic grouping algorithm. The size of the similarity matrix grows quadratically with the number of input points, which limits the size of point clouds that can be processed. In order to simultaneously segment instances and semantic object categories, several methods\replaced[id=R2]{~\citep{Wang2019AssociativelySI,Pham2019JSIS3DJS,Elich20193DBEVISBI,Lahoud20193DIS,han2020occuseg,He2020LearningAM,Zhang2019PointCI}}{\mbox{~\citep{Wang2019AssociativelySI,Pham2019JSIS3DJS,Lahoud20193DIS,Elich20193DBEVISBI,Zhang2019PointCI,He2020LearningAM,han2020occuseg}}} employ parallel decoder branches for the point classification and the discriminative instance embedding. There are a number of variants of this general idea, e.g.,~\cite{Wang2019AssociativelySI} introduce a linking module between the two decoder branches to exploit synergies between semantic and instance segmentation. \cite{Pham2019JSIS3DJS} employ a multi-value conditional random field model to jointly optimise instance and semantic labels. \cite{Lahoud20193DIS} estimate vectors pointing to the potential instance centers, as in the generalised Hough transform, to support the subsequent clustering step. 
In addition to the 3D offset vector, OccuSeg~\citep{han2020occuseg} also learns occupancy signals, which can guide the subsequent graph-based clustering towards better instance segmentation. \cite{Elich20193DBEVISBI} integrate 2D birds-eye-view information into a network for joint 3D semantic and instance segmentation, in order to better exploit global context. \cite{Zhang2019PointCI} introduce a probabilistic embedding instead of a deterministic one, and also propose a new loss function for the clustering step, with which they achieved good performance on the PartNet dataset~\citep{Mo2019PartNetAL}. In order to mitigate imbalances in the data, which tend to harm the instance segmentation for rare categories,~\cite{He2020LearningAM} propose a memory-augmented network to memorise representative patterns. 

Recently, a number of studies have applied the bottom-up approach to outdoor dataset~\citep{milioto2020lidar,hong2021lidar,zhou2021panoptic,zhao2021technical,li2021cpseg}. In general, these are variants of the two-branch architecture described above. For instance,~\cite{milioto2020lidar} and~\cite{li2021cpseg} used spherical projection to implement a real-time, panoptic segmentation algorithm for the autonomous driving setting, while Panoptic-PolarNet~\citep{zhou2021panoptic} used a polar bird's eye view. Their instance branch directly regresses the instance's center. DS-Net~\citep{hong2021lidar} utilizes a dynamic shift module that can automatically adjust the kernel function to different point densities and instance sizes. All these works target autonomous driving scenarios with sparse, vehicle-mounted panoramic LiDAR, in particular the popular SemanticKITTI~\citep{behley2019iccv} and nuScenes~\citep{nuscenes2019} datasets. To our knowledge, none of them has been tuned to panoptic segmentation of dense, mobile-mapping type point clouds, neither have they been evaluated on such datasets, most prominently NPM3D~\citep{roynard2017parislille3d}.

Compared to the 2D image case, 3D instance segmentation is not as mature, and there is still ample room for improvement. Even though discriminative per-point embedding features followed by clustering have emerged as the mainstream approach, there still are a number of open questions. One elementary issue is that object instances in 3D scans vary greatly in size and point density, leading to over- and/or under-segmentation when using fixed clustering parameters \replaced[id=R2]{(e.g., the bandwidth $Bw$ in the case of mean-shift~\citep{Comaniciu2002Mean})}{\mbox{~\citep[e.g., the bandwidth $Bw$ in the case of mean-shift,][]{Comaniciu2002Mean}}}. Therefore, there recently have been attempts to improve the clustering step\replaced[id=R2]{~\citep{Engelmann20CVPR,jiang2020pointgroup,jiang2020end,he2021dyco3d,chen2021hierarchical,liang2021instance}}{\mbox{~\citep{jiang2020pointgroup,Engelmann20CVPR,jiang2020end,he2021dyco3d,chen2021hierarchical,liang2021instance}}}. The process usually starts by obtaining a sufficient number of instance candidates, which are then subject to various merging, pruning and de-duplicating operations to obtain the final instances. E.g., PointGroup~\citep{jiang2020pointgroup} merges clusters based on both their original and embedded coordinates to increase variety, scores the resulting instance candidates with a learned ScoreNet, then runs non-maximum suppression to prune overlapping candidates. 3D-MPA~\citep{Engelmann20CVPR} and HAIS~\citep{chen2021hierarchical} favour aggregating candidates over non-maxima suppression. 3D-MPA generates multiple instance candidates by randomly sampling from the predicted instance centers, then constructs a graph convolutional network to allow for information exchange between adjacent candidates and find the best aggregation. Like the previously discussed methods, HAIS~\citep{chen2021hierarchical} clusters points into preliminary instance candidates based on semantics and location, then each candidate attempts to absorb nearby ones with a dynamic bandwidth that is proportional to its initial size. Finally, a neural network again acts as a score function to select instances. DyCo3D~\citep{he2021dyco3d} and SSTNet~\citep{liang2021instance} also follow similar strategies. After obtaining the preliminary instance candidates, DyCo3D dynamically generates a filter to predict the binary instance mask for each candidate. Instead of directly clustering the points to generate preliminary instance candidates, SSTNet first generates geometrically homogeneous super-points, which are then linked into a tree. That tree serves as a basis for divisive grouping to find instance candidates, which are again refined by pruning and a scoring network. The SoftGroup network~\citep{Vu2022SoftGroup} also made improvements based on HAIS. In the clustering phase, each point is not first assigned one semantic class then clustered. Instead, a point may be associated with multiple possible semantic classes and, as a result, may belong to clusters in multiple classes. By doing so, the effects of semantic classification errors on subsequent instance segmentation will be mitigated. ASNet~\citep{jiang2020end} converts the point clustering problem into an assignment problem, i.e., it samples $k$ instance candidates and predicts, for each point, which of those candidates it belongs to. Note that at training time this involves bipartite matching~\citep{kuhn1955hungarian} between the predicted and true instance centers, before one can compute their cross-entropy. The method also features a suppression module that learns to remove redundant candidates. 

%-------------------------------------------------------------------------
\subsection{Other strategies}
\label{subSec:OtherStrategies}
%-------------------------------------------------------------------------

Panoster~\citep{gasperini2021panoster} outputs instance IDs directly, without having to sample or cluster points. The trick is to construct a softmax-based confusion matrix between real and predicted instances, and then use an impurity loss to boost the correlation between real and predicted values, together with a fragmentation loss to discourage instances with few points. ICM-3D~\citep{chu2021icm} reformulates the instance segmentation problem into a classification problem inspired by SOLO~\citep{wang2021solo}, an effective 2D instance segmentation method. It eliminates additional clustering step to extract candidates, and avoids using non-maximum suppression to remove duplicate candidates. However, so far, this strategy is not competitive with clustering-based approaches in terms of segmentation quality.

Recently, the success of transformers for 2D segmentation has inspired a few works that employ transformers for 3D instance segmentation\replaced[id=R2]{~\citep{Liu20223D-QueryIS,Sun2022Superpoint,Schult2023Mask3D}}{\mbox{~\citep{Schult2023Mask3D,Liu20223D-QueryIS,Sun2022Superpoint}}}. They establish a query decoder that directly generates semantic classes, scores, and instance mask predictions. The predicted masks and ground truth masks are automatically associated using bipartite matching, enabling end-to-end  training. Transformers achieve competitive performance, but have high computational overhead due to the complexity of the attention mechanism and the associated increase in the number of trainable parameters. How to more efficiently use the attention mechanism and find a good trade-off between efficiency and accuracy may be an interesting further direction for 3D instance segmentation.

%-------------------------------------------------------------------------
\section{Evaluation metrics and datasets}
\label{Sec:EvaluationMetricsAndDatasets}
%-------------------------------------------------------------------------

%-------------------------------------------------------------------------
\subsection{Evaluation metrics}
\label{subSec:EvaluationMetrics}
%-------------------------------------------------------------------------
Depending on the application, it may be more important to retrieve scene semantics or to correctly delineate instances. Different evaluation metrics are suited for the two subtasks of semantic segmentation and instance segmentation, and for the joint task of panoptic segmentation. For a comprehensive evaluation, all performance metrics are computed in the following experiments.

\noindent\textbf{Evaluation of semantic segmentation}: Common metrics for semantic segmentation include overall accuracy (oAcc), which is however biased towards classes with many points;
%mean accuracy (mAcc) over all categories,
and the mean Intersection-over-Union (mIoU) over all categories. With $N$ the number of total points, $C$ the number of categories, $\text{TP}_i$ the number of points correctly assigned to semantic category $i$, $\text{GT}_i$ the number of points with true semantic label $i$, and $\text{PRE}_i$ the number of points with predicted semantic label $i$, the metrics are defined as:
\begin{equation}\label{Eq:OverallAccuracy}
	\text{oAcc}=\frac{\sum_{i=1}^{C}\text{TP}_i}{N}\;,\quad
	\text{mIoU}=\frac{1}{C}\sum_{i=1}^{C}\frac{\text{TP}_i}{\text{GT}_i+\text{PRE}_i-\text{TP}_i}\;.
\end{equation}

%\begin{equation}\label{Eq:MeanAccuracy}
%	\text{mAcc}=\frac{1}{C}\sum_{i=1}^{C}\frac{\text{TP}_i}{\text{GT}_i},
%\end{equation}

\noindent\textbf{Evaluation of instance segmentation}: To evaluate instance segmentation, one first removes all points assigned to ``stuff'' classes that do not have distinct instances. For the remaining ``things'' classes, common metrics are mean coverage (mCov), mean weighted coverage (mWCov), mean precision (mPrec), mean recall (mRec) and F1-score. First, one collects all points that share the same predicted instance label into an instance prediction. The semantic category of that instance is determined by majority voting and assigned to all its points. In each category $i$, a set of ground truth semantic instances $\{I_j^\text{gt},j\in\{1,...,N_\text{gt}^i\}\}$, and a set of predicted instances $\{I_k^\text{pre},k\in\{1,...,N_\text{pre}^i\}\}$ are given. Moreover, an operator that compares an instance from one set with all instances in the other and returns the highest intersection-over-union score is defined as:
\begin{comment}
\begin{equation}
\begin{split}
    \text{maxIoU}(I_j^\text{gt}) = \max_{k=1}^{N_\text{pre}^i} \big(\text{IoU}(I_j^\text{gt},I_k^\text{pre}) \big)\\
     \text{maxIoU}(I_k^\text{pre}) = \max_{j=1}^{N_\text{gt}^i} \big(\text{IoU}(I_j^\text{gt},I_k^\text{pre}) \big)
    \end{split}
\end{equation}
\end{comment}
\begin{equation}
    \text{maxIoU}(I_j^\text{gt}) = \max_{k=1}^{N_\text{pre}^i} \big(\text{IoU}(I_j^\text{gt},I_k^\text{pre}) \big)\;,\quad
     \text{maxIoU}(I_k^\text{pre}) = \max_{j=1}^{N_\text{gt}^i} \big(\text{IoU}(I_j^\text{gt},I_k^\text{pre}) \big)\;.
\end{equation}%
The coverage in category $i$ is then defined as:
\begin{equation}\label{Eq:Cov}
	\text{Cov}_i=\frac{1}{N_\text{gt}^i}\sum_{j=1}^{N_\text{gt}^i}\text{maxIoU}(I_j^\text{gt})\;,\quad
	\text{WCov}_i=\frac{1}{N_\text{gt}^i}\sum_{j=1}^{N_\text{gt}^i}{w_j}\cdot\text{maxIoU}(I_j^\text{gt})\;,
\end{equation}
where $\lvert{I_j^\text{gt}}\rvert$ is the number of points in instance $I_j^\text{gt}$, and the weight 
$w_j = \lvert{I_j^\text{gt}}\rvert/\sum_{m=1}^{N_\text{gt}^i}{\lvert{I_m^\text{gt}}\rvert}$. For the further scores, one first discards all predicted instances that have $\text{maxIoU}(I_k^\text{pre})<0.5$, to obtain the set of valid predictions $\{I_l^\text{val},l\in[1,\dots,N_\text{val}^i]\}$. With the number $N_\text{val}^i$ of valid predictions, the precision and recall in category $i$ are defined as 
\begin{equation}\label{Eq:PrecRec}
	\text{Prec}_i=\frac{N_\text{val}^i}{N_\text{pre}^i}\;,\quad\text{Rec}_i=\frac{N_\text{val}^i}{N_\text{gt}^i}\;.
\end{equation}
The overall metrics are then found by computing the means across all categories:
\begin{align}\label{Eq:mCov_mWCov}
	&\text{mCov}=\frac{1}{C}\sum_{i=1}^{C}\text{Cov}_i\;, &\text{mWCov}=\frac{1}{C}\sum_{i=1}^{C}\text{WCov}_i\;,\\
	&\text{mPrec}=\frac{1}{C}\sum_{i=1}^{C}\text{Prec}_i\;, &\text{mRec}=\frac{1}{C}\sum_{i=1}^{C}\text{Rec}_i\;,
\end{align}
and the F1-score is calculated in the usual manner as
\begin{equation}\label{Eq:F1-score}
	\text{F1}=\frac{2\cdot\text{mPrec}\cdot\text{mRec}}{\text{mPrec}+\text{mRec}}\;.
\end{equation}

\noindent\textbf{Evaluation of panoptic segmentation}: For panoptic segmentation, it is natural to also adopt the metrics proposed for the 2D image version of the task, namely panoptic quality (PQ), segmentation quality (SQ), recognition quality (RQ), and PQ\dag~\citep{Kirillov2019Panoptic,porzi2019seamless}. Again one first computes the scores separately per semantic category $i$, regarding all points as a single, big instance for the ``stuff'' categories. For the segmentation quality, one accumulates the IoU of all true positive instances:
\begin{equation}\label{Eq:SQ}
	\text{SQ}_i=\frac{1}{N_\text{val}^i}\sum_{l=1}^{N_\text{val}^i} \text{maxIoU}(I_l^\text{val})\;.
\end{equation}
\begin{comment}
where
\begin{equation}\label{Eq:maxIoUIval}
	\text{maxIoU}(I_l^\text{val})=\max_{j=1}^{N_\text{gt}^i} \big(\text{IoU}(I_j^\text{gt},I_l^\text{val}) \big)\;,
\end{equation}
\end{comment}
The recognition quality is simply the per-category F1-score,
\begin{equation}\label{Eq:RQ}
	\text{RQ}_i=\text{F1}_i=\frac{2\cdot\text{Prec}_i\cdot\text{Rec}_i}{\text{Prec}_i+\text{Rec}_i}\;,
\end{equation}
and the panoptic quality combines the two previous metrics into a single number:
\begin{equation}\label{Eq:PQ}
	\text{PQ}_i=\text{SQ}_i\cdot\text{RQ}_i\;.
\end{equation}
Note that for each ``stuff'' class there can be at most 1 ground truth instance and 1 predicted instance in a point cloud. Hence the SQ, RQ and PQ metrics are not properly defined for ``stuff'' classes with IoU below 0.5, or penalise them very harshly if set to 0. \cite{porzi2019seamless} therefore proposed an improved metric PQ\dag, which uses PQ for ``things'' classes, but replaces it with the simple IoU for ``stuff'' classes.
\begin{equation}\label{Eq:PQ_star}
{\text{PQ}\dag}_i=
\begin{cases}
\text{IoU}(I^\text{gt},I^\text{pre}) & \text{if class $i$ is ``stuff''}\\
\text{PQ}_i & \text{otherwise .}
\end{cases}
\end{equation}
In case the ``stuff'' classes have sufficient overlap to be valid, the difference between PQ and PQ\dag\ is small, which is always the case in our experiments. Nevertheless both are quoted for completeness.
 
As before, the final panoptic metrics are calculated by averaging the respective per-category scores over all semantic categories:
\begin{equation}\label{Eq:meanPano}
	\text{SQ}=\frac{1}{C}\sum_{i=1}^{C}\text{SQ}_i\;,\quad\text{RQ}=\frac{1}{C}\sum_{i=1}^{C}\text{RQ}_i\;,\quad
	\text{PQ}=\frac{1}{C}\sum_{i=1}^{C}\text{PQ}_i\;,\quad\text{PQ\dag}=\frac{1}{C}\sum_{i=1}^{C}\text{PQ\dag}_i\;.
\end{equation}

%-------------------------------------------------------------------------
\subsection{Dataset}
\label{subSec:Dataset}
%-------------------------------------------------------------------------

This section briefly introduces a number of popular, publicly available 3D point cloud datasets suitable for training and testing panoptic segmentation. A distinction is made between indoor and outdoor datasets, which have rather different characteristics in terms of both scene content and sensor parameters.

%-------------------------------------------------------------------------
\subsubsection{Outdoor datasets}
\label{subSec:Outdoordataset}
%-------------------------------------------------------------------------

\noindent\textbf{NPM3D}~\citep{roynard2017parislille3d} is a public benchmark for point cloud semantic segmentation, with 10 classes including: ground, building, pole (road sign and traffic light), bollard, trash can, barrier, pedestrian, car, natural (vegetation) and unclassified. Results are evaluated only w.r.t.\ 9 classes, disregarding the ``unclassified'' label. The data has been captured with a mapping-grade mobile laser scanning system in different cities in France. There are 4 regions designated for training, all captured in Paris and Lille; and 3 regions for testing, captured in Dijon and Ajaccio.
The standard 10-class version described above has actually been derived from a more fine-grained version of the dataset by keeping only the most frequent labels. The original annotations feature 50 different semantic classes (most of which are very rare), and also individual object instance labels for the training regions. For panoptic segmentation, a new version has been generated that still uses the 10 semantic category labels listed above, but also includes instance labels (available at \replaced[id=R2]{\url{https://doi.org/10.5281/zenodo.8188390}}{\url{https://polybox.ethz.ch/index.php/s/vLepzlJtIofdTtE}}). The classes ground, building and barrier are considered ``stuff'' and are not separated into instances. As no instance labels are available for the 3 test regions, our version for panoptic (or pure instance) segmentation only contains 4 different regions from Paris and Lille. Instead of a fixed training/test split all experiments therefore use 4-fold cross-validation\replaced[id=R2]{}{\ }.

Other outdoor datasets for panoptic segmentation exist, which target the rather specific situation of autonomous driving. They have a very different sensing geometry, namely sequences of individual, sparse 32-beam panorama scans, and are thus not as suitable for our target application of mobile mapping. \textbf{SemanticKITTI}~\citep{behley2019iccv, behley2021ijrr} provides annotations for 22 scenes with 28 semantic classes, for a total of 23'201 scans for training, and 20'351 scans for testing. \textbf{nuScenes}~\citep{nuscenes2019,fong2021panoptic} is a similar dataset of much larger scale featuring 16 classes (10 ``things'' and 6 ``stuff''), with 1000 scenes collected in Singapore and Boston, totalling 300k scans.

%-------------------------------------------------------------------------
\subsubsection{Indoor datasets}
\label{subSec:Indoordataset}
%-------------------------------------------------------------------------

\noindent\textbf{S3DIS}~\citep{Iro20163DSemantic}, the ``Stanford 3D Indoor Scene Dataset'', contains 6 large-scale indoor areas with a total of 271 rooms. Data were collected in three different office/university buildings using the Matterport scanner.\footnote{\url{http://matterport.com/}}
In total the colored 3D point clouds have $\approx$696M points. Each point is annotated with one out of 13 semantic categories and an instance ID. There are two common experimental protocols: one uses area 5 as a fixed test set, the other one employs 6-fold cross-validation. S3DIS is perhaps the most widely used indoor dataset for instance segmentation, and among the datasets that provide instance annotations arguably also the most relevant one for mobile mapping in terms of data quality. Hence, it is used for complementary indoor experiments, see the~\ref{Sec:ExperimentsonS3DISdataset}.

Other indoor datasets for panoptic segmentation include \noindent\textbf{ScanNet}~\citep{dai2017scannet}, a collection of labeled voxels (rather than points). The current version, ScanNet v2, has 1513 scans, annotated with 20 semantic classes (18 ``things'' and 2 ``stuff'') and instance IDs. There is a prescribed split into 1201 training scans, 312 validation scans and 100 test scans (with private labels). \noindent\textbf{SceneNN}~\citep{Hua2016SceneNN} is a smaller indoor RGB-D dataset of 100 scenes. Of those, 76 scenes have been annotated with 40 semantic categories, split into 56 training scenes and 20 test scenes.

\begin{sidewaystable}[htbp]
\renewcommand\arraystretch{2}
\caption{Summary of existing work on bottom-up panoptic point cloud segmentation, underlying methods and datasets. Rows sorted in chronological order.}
\centering
\resizebox{\textwidth}{!}{\begin{tabular}{c c cccc ccccc}
		\toprule
		\multirow{2}{*}{Methods} & \multirow{2}{*}{Backbone} & \multicolumn{4}{c}{Features extracted by semantic and instance segmentation branch} & \multicolumn{5}{c}{Dataset} \\
		\cline{3-11} 
		& & \multicolumn{1}{c}{Semantic features} & \multicolumn{1}{c}{Embedding features} & \multicolumn{1}{c}{Offsets} & Others & \multicolumn{1}{c}{S3DIS} & \multicolumn{1}{c}{Scannet} & \multicolumn{1}{c}{SemanticKITTI} &
		\multicolumn{1}{c}{nuScenes} & others\\ \midrule
		\multicolumn{1}{l}{ASIS~\citep{Wang2019AssociativelySI}} & \multicolumn{1}{l}{PointNet~\citep{Qi2017PointNetDL}/PointNet++~\citep{Qi2017PointNetDH}} & \multicolumn{1}{c}{\checkmark} & \multicolumn{1}{c}{\checkmark} & \multicolumn{1}{c}{} & & \multicolumn{1}{c}{\checkmark} & \multicolumn{1}{c}{} & \multicolumn{1}{c}{} & \multicolumn{1}{c}{} & \multicolumn{1}{l}{ShapeNet~\citep{Yi2016Scalable}}    \\ 
		\multicolumn{1}{l}{JSIS3D~\citep{Pham2019JSIS3DJS}} &        \multicolumn{1}{l}{PointNet~\citep{Qi2017PointNetDL}}  & \multicolumn{1}{c}{\checkmark} & \multicolumn{1}{c}{\checkmark} & \multicolumn{1}{c}{} & & \multicolumn{1}{c}{\checkmark} & \multicolumn{1}{c}{} & \multicolumn{1}{c}{} & \multicolumn{1}{c}{} &  \multicolumn{1}{l}{SceneNN~\citep{Hua2016SceneNN}} \\ 
		\multicolumn{1}{l}{MTML~\citep{Lahoud20193DIS}}& \multicolumn{1}{l}{3D convolution network based on the SSCNet~\citep{song2017semantic} } & \multicolumn{1}{c}{\checkmark} & \multicolumn{1}{c}{\checkmark} & \multicolumn{1}{c}{} & \multicolumn{1}{l}{Direction embedding features} & \multicolumn{1}{c}{} & \multicolumn{1}{c}{\checkmark} & \multicolumn{1}{c}{} &
		\multicolumn{1}{c}{} & \\ 
		\multicolumn{1}{l}{MPNet~\citep{He2020LearningAM}}&
		\multicolumn{1}{l}{PointNet++~\citep{Qi2017PointNetDH}} & \multicolumn{1}{c}{\checkmark} & \multicolumn{1}{c}{\checkmark} & \multicolumn{1}{c}{} & & \multicolumn{1}{c}{\checkmark} & \multicolumn{1}{c}{\checkmark} & \multicolumn{1}{c}{} & \multicolumn{1}{c}{} &  \multicolumn{1}{l}{PartNet~\cite{Mo2019PartNetAL} } \\ 
		\multicolumn{1}{l}{3D-MPA~\citep{Engelmann20CVPR}}&  \multicolumn{1}{l}{Minkowski CNN~\citep{choy20194d}} & \multicolumn{1}{c}{\checkmark} & \multicolumn{1}{c}{} & \multicolumn{1}{c}{\checkmark} & & \multicolumn{1}{c}{\checkmark} & \multicolumn{1}{c}{\checkmark} & \multicolumn{1}{c}{} & \multicolumn{1}{c}{} & \\ 
		\multicolumn{1}{l}{PointGroup~\citep{jiang2020pointgroup}}& \multicolumn{1}{l}{Submanifold sparse U-Net~\citep{Graham20183DSemantic}} & \multicolumn{1}{c}{\checkmark} & \multicolumn{1}{c}{} & \multicolumn{1}{c}{\checkmark} & & \multicolumn{1}{c}{\checkmark} & \multicolumn{1}{c}{\checkmark} & \multicolumn{1}{c}{} & \multicolumn{1}{c}{} & \\ 
		\multicolumn{1}{l}{OccuSeg~\citep{han2020occuseg}}& \multicolumn{1}{l}{Submanifold sparse U-Net~\citep{Graham20183DSemantic} } & \multicolumn{1}{c}{\checkmark} & \multicolumn{1}{c}{\checkmark} & \multicolumn{1}{c}{\checkmark} & \multicolumn{1}{l}{Occupancy features} & \multicolumn{1}{c}{\checkmark} & \multicolumn{1}{c}{\checkmark} & \multicolumn{1}{c}{} & \multicolumn{1}{c}{} &  \multicolumn{1}{l}{SceneNN~\citep{Hua2016SceneNN}} \\ 
		\multicolumn{1}{l}{ASNet~\citep{jiang2020end}}& \multicolumn{1}{l}{PointNet~\citep{Qi2017PointNetDL}/PointNet++~\citep{Qi2017PointNetDH} } & \multicolumn{1}{c}{\checkmark} & \multicolumn{1}{c}{\checkmark} & \multicolumn{1}{c}{} & \multicolumn{1}{l}{3D instance centroid and centroid distance}  & \multicolumn{1}{c}{\checkmark} & \multicolumn{1}{c}{} & \multicolumn{1}{c}{} & \multicolumn{1}{c}{} &  \multicolumn{1}{l}{SceneNN~\citep{Hua2016SceneNN}} \\
		\multicolumn{1}{l}{DS-Net~\citep{hong2021lidar}}&  \multicolumn{1}{l}{Cylinder 3D~\citep{zhou2020cylinder3d} } & \multicolumn{1}{c}{\checkmark} & \multicolumn{1}{c}{} & \multicolumn{1}{c}{\checkmark} & & \multicolumn{1}{c}{} & \multicolumn{1}{c}{} & \multicolumn{1}{c}{\checkmark} & \multicolumn{1}{c}{\checkmark} & \\
		\multicolumn{1}{l}{Panoster~\citep{gasperini2021panoster}} & \multicolumn{1}{l}{KPConv~\citep{Thomas2019KPConvFA}/SalsaNext~\citep{cortinhal2020salsanext}} & \multicolumn{1}{c}{\checkmark} & \multicolumn{1}{c}{} & \multicolumn{1}{c}{} & & \multicolumn{1}{c}{} & \multicolumn{1}{c}{} & \multicolumn{1}{c}{\checkmark} &  
		\multicolumn{1}{c}{} & \\ 
		\multicolumn{1}{l}{DyCo3D~\citep{he2021dyco3d}}& \multicolumn{1}{l}{Submanifold sparse U-Net~\citep{Graham20183DSemantic}} & \multicolumn{1}{c}{\checkmark} & \multicolumn{1}{c}{} & \multicolumn{1}{c}{\checkmark} & & \multicolumn{1}{c}{\checkmark} & \multicolumn{1}{c}{\checkmark} & \multicolumn{1}{c}{} & \multicolumn{1}{c}{} & \\
		\multicolumn{1}{l}{~\cite{Zhang2019PointCI}}& \multicolumn{1}{l}{PointNet++~\citep{Qi2017PointNetDH}} & \multicolumn{1}{c}{\checkmark}   & \multicolumn{1}{c}{}   & \multicolumn{1}{c}{}  & \multicolumn{1}{l}{Probabilistic embedding features}  & \multicolumn{1}{c}{} & \multicolumn{1}{c}{\checkmark} & \multicolumn{1}{c}{} & 
		\multicolumn{1}{c}{} &\multicolumn{1}{l}{PartNet~\cite{Mo2019PartNetAL}} \\
		\multicolumn{1}{l}{HAIS~\citep{chen2021hierarchical}}& \multicolumn{1}{l}{Submanifold sparse U-Net~\citep{Graham20183DSemantic}}  & \multicolumn{1}{c}{\checkmark} & \multicolumn{1}{c}{} & \multicolumn{1}{c}{\checkmark} &  & \multicolumn{1}{c}{\checkmark} & \multicolumn{1}{c}{\checkmark} & \multicolumn{1}{c}{} &  \multicolumn{1}{c}{} &   \\ 
		\multicolumn{1}{l}{SSTNet~\citep{liang2021instance}}&    \multicolumn{1}{l}{Submanifold sparse U-Net~\citep{Graham20183DSemantic}}  & \multicolumn{1}{c}{\checkmark} & \multicolumn{1}{c}{}  & \multicolumn{1}{c}{\checkmark} &  & \multicolumn{1}{c}{\checkmark} & \multicolumn{1}{c}{\checkmark}  & \multicolumn{1}{c}{}  &  \multicolumn{1}{c}{}  &   \\
		\multicolumn{1}{l}{SoftGroup~\citep{Vu2022SoftGroup}}&    \multicolumn{1}{l}{Submanifold sparse U-Net~\citep{Graham20183DSemantic}}  & \multicolumn{1}{c}{\checkmark} & \multicolumn{1}{c}{}  & \multicolumn{1}{c}{\checkmark} &  & \multicolumn{1}{c}{\checkmark} & \multicolumn{1}{c}{\checkmark}  & \multicolumn{1}{c}{}  &  \multicolumn{1}{c}{}  &   \\\bottomrule
\end{tabular}}
\label{table:summarizesmethods}
\end{sidewaystable}

A number of recent papers have studied panoptic point cloud segmentation. Among those, the present review focuses on the most successful school, namely bottom-up approaches that directly assign instance labels to points and do not rely on dedicated bounding box detectors. Table~\ref{table:summarizesmethods} presents a summary. According to the motivation of the present paper (c.f.\ Section~\ref{Sec:Introduction}), the Table~\ref{table:summarizesmethods} looks at three fundamental attributes: (1)  the 3D backbone architecture used for feature extraction, (2) the type of feature representation used for instance clustering, and (3) the dataset(s) used for evaluation. Methods based on 2D projection and subsequent feature extraction with 2D backbones have been omitted, as they have only been used for the autonomous driving setting, where a natural projection exists from 3D points to 2.5D scanner coordinates. One can readily see in Table~\ref{table:summarizesmethods} that PointNet++ and sparse voxel CNNs (Submanifold Sparse U-Net and Minkowski Engine) are the two most popular backbones. Since panoptic segmentation includes semantic segmentation, all methods extract a set of features optimised for semantic classification, which serve as input for the corresponding classifier. Additionally, most methods extract some dedicated representation for instance segmentation, most often either another set of features, optimised to discriminate instances, or an offset vector from each point to its associated instance center. Indoor panoptic segmentation is commonly trained and tested on S3DIS and/or ScanNet, SceneNN and other (synthetic) datasets are less widespread. Very few works have addressed outdoor settings, and those which have are restricted to the sparse, panoramic scans of the autonomous driving scenario. Instance segmentation of dense, full 3D mobile mapping data does not seem to have been attempted with modern, neural network-based methods.

%-------------------------------------------------------------------------
\section{Experiments}
\label{Sec:Experiments}
%-------------------------------------------------------------------------

%-------------------------------------------------------------------------
\subsection{Panoptic segmentation pipeline}
\label{subSec:ProposedPipeline}
%-------------------------------------------------------------------------
\begin{figure}[htbp]
	\centering
	\includegraphics[width=\linewidth]{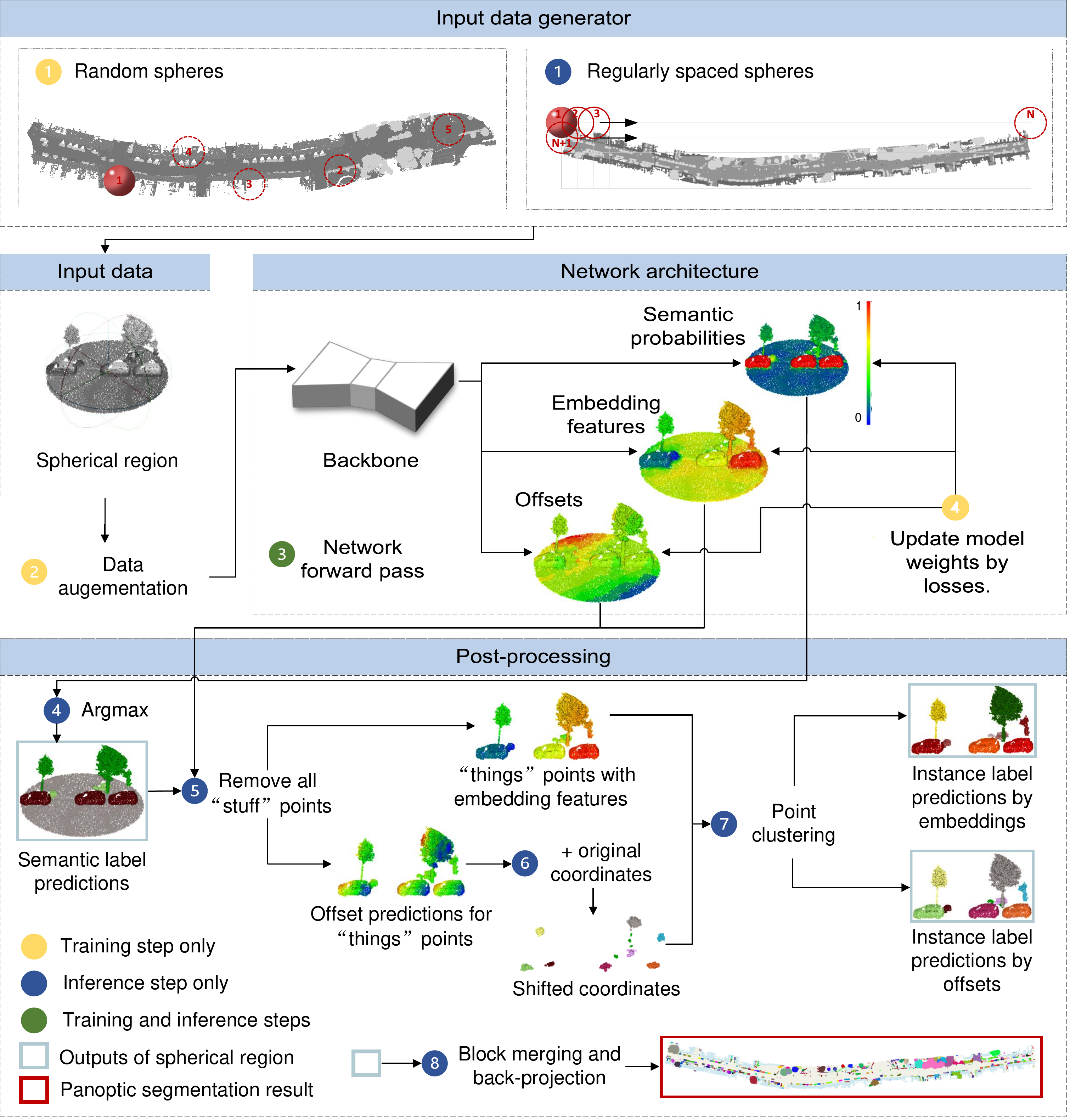}\\
	\vspace{0cm}
	\caption{\added[id=R2]{Proposed panoptic segmentation pipeline for mobile mapping point clouds.}}
	\label{Fig:pipeline}
\end{figure}

For the experiments, our study adopts the bottom-up instance segmentation strategy, due to its superior performance in recent experimental studies\replaced[id=R2]{~\citep{Engelmann20CVPR,jiang2020pointgroup,jiang2020end,he2021dyco3d,chen2021hierarchical,liang2021instance}}{\mbox{~\citep{jiang2020pointgroup,Engelmann20CVPR,jiang2020end,he2021dyco3d,chen2021hierarchical,liang2021instance}}}. A detailed graphical illustration is given in Figure~\ref{Fig:pipeline}. Notations and descriptions of all parameters appear in this paper can be found in Table~\ref{table:allParametersNotationAndDescription}.

\renewcommand\arraystretch{0.8}
\begin{table}[htbp]
\caption{Notation and description of parameters.}
\vspace{0.3cm}
%\resizebox{\textwidth}{!}{
\centering
\begin{tabular}{lll}
\hline
Category & Parameter & Description  \\ \hline
\multirow{5}{*}{Input}      & $d$       & voxel grid size for down-sampling \\
& $R$       & radius of input spheres \\
& $s$       & sphere sampling stride during inference \\
& $k$       & fixed number of points/sphere if using PointNet++ \smallskip \\
& $K$   & \begin{tabular}[c]{@{}l@{}}dimension of input features for vector $F$: \\K=3 $\dots$ input features are relative $(X,Y,Z)$ coordinates,\\ K=4 $\dots$ relative $(X,Y,Z)$ and original $Z$,\\ K=6 $\dots$ relative $(X,Y,Z)$ and RGB,\\ K=7 $\dots$ relative $(X,Y,Z)$ and original $Z$ and RGB.\end{tabular} \\ \hline
\multirow{3}{*}{Training}   & $W_{e}$   & weight of embedding loss \\
& $W_{o}$   & weight of offset loss \\
& $W_{r}$   & weight of regulariser \\ \hline
\multirow{4}{*}{Thresholds} & $Th_d $   & threshold for instance clustering  \\
& $Th_n$    & minimum point number for a valid instance \\
& $Th_{bm}$ & threshold for block merging  \\
& $Bw$      & mean-shift kernel bandwidth  \\ \hline
\end{tabular}
\label{table:allParametersNotationAndDescription}
\end{table}

%-------------------------------------------------------------------------
\subsubsection{Input}
\label{subsubSec:InputData}
%-------------------------------------------------------------------------

Mobile mapping point clouds of realistic extent and density are far too large for processing on current graphics hardware. To reduce the excessive point density in the near field of the scanner, the raw data are preprocessed with voxel-grid downsampling, keeping only one (randomly selected) point per $d\times d\times d$ voxel. Even after downsampling it is not possible to process larger regions in one piece. Hence, one must in practice work with local neighborhoods, potentially sacrificing some long-range context. In this study spherical neighborhoods of a fixed radius have been used. The set of points inside the sphere, denoted as $\mathbb{P}$, is transformed to relative coordinates w.r.t.\ the center point. If other per-point information shall be used (e.g., RGB color values or height in absolute scene coordinates), they are appended to the relative coordinates to form a $K$-dimensional input vector $F$ per point. For example, with $K\!=\!6$ the vector $F$ contains the relative location of a 3D point and the RGB information. All possible feature combinations and hyperparamters are listed in Table~\ref{table:allParametersNotationAndDescription}. The features of the point set $\mathbb{P}$ are denoted $\mathbf{F}=\left\{F_{i}\right\}\in\mathbb{R}^{N\times{K}}$, with $N$ the number of points.

The training stage randomly samples the spheres' center locations in the point cloud. Additionally, data augmentation is performed by random scaling and rotation around the vertical axis, as well as adding Gaussian noise to the point coordinates. When working with the PointNet++ backbone, which requires a fixed number of points per input sphere, random duplication or removal of points are applied to reach that number. At inference time a regular grid of overlapping spheres is used, with fixed stride $s$ along the three coordinate axes. 

%-------------------------------------------------------------------------
\subsubsection{Network architecture}
\label{subsubSec:NetworkArchitecture}
%-------------------------------------------------------------------------

Out of the numerous neural network architectures that are nowadays available to encode 3D point coordinates (and associated input features) into a predictive latent representation, our study evaluates three representative ones as backbones for the experimental pipeline, see Table~\ref{table:summarizesmethods}: PointNet++, denoted as PN$(x)$, is a popular and widely used architecture, and arguably the best-performing representative from the first generation of point cloud networks, based on per-point multi-layer perceptrons (MLPs) with shared weights, rather than convolutions. Sparse CNN (as implemented in the Minkowski Engine), denoted as MC$(x)$, is a recent high-performance variant of standard, discrete convolution on sparse 3D voxel grids. And KPConv, denoted as KP$(x)$, is the perhaps most elaborate approximation of convolution in continuous space, sparsified by only sampling the (continuous) output at the original point locations. In our pipeline the backbones are interchangeable and operate as a black box module that turns the input into a per-point feature representation of fixed size (channel depth). That representation is further fed into two parallel heads responsible for semantic segmentation and instance segmentation (Figure~\ref{Fig:pipeline}). The semantic segmentation head $H^1(\cdot)$ predicts, for every point, the membership probabilities $\mathbf{CP}$ for all semantic categories. For the instance segmentation head there are again two variants: $H^2(\cdot)$ transforms the input representation into an embedding feature space $\mathbf{EF}$ such that points on the same instance form compact clusters; whereas $H^3(\cdot)$ instead estimates per-point 3D offset vectors $\mathbf{O}$ pointing to the center of its associated instance. All three heads are three-layer MLPs with the same structure. Their first two layers are linear transformations with the same number of input and output channels, followed by Batch Normalization~\citep{ioffe2015batch} and LeakyReLU activations~\citep{maas2013rectifier}. The output size of the third layer differs between the three heads: for $H^1$ it is the number $C$ of semantic categories, for $H^2$ it is the (tunable) dimension of the embedding space, and for $H^3$ it is a 3D displacement vector.

%-------------------------------------------------------------------------
\subsubsection{Loss functions}
\label{subsubSec:LossFunctions}
%-------------------------------------------------------------------------

\noindent\textbf{Semantic segmentation head}: The output of this network branch are class probabilities $\mathbf{CP}=\left\{CP_{i}\right\}\in\mathbb{R}^{N\times{C}}$ for $N$ points and $C$ semantic classes, with $i\in\left\{1,...,N\right\}$. The corresponding loss function $L_{s}$ is the standard cross-entropy between the rows of $\mathbf{CP}$ and the ground truth labels $y$:
\begin{equation}\label{Eq:Cross-entropyLoss}
	L_{s}=-\frac{1}{N}\sum_{i=1}^{N}\sum_{j=1}^{C}y_{ij}log({{CP}_{ij}})\added{.}
\end{equation}

\noindent\textbf{Instance embedding head}: The output of this branch $\mathbf{EF}=\left\{EF_{i}\right\}\in\mathbb{R}^{N\times{T}}$ is an embedding of the points in a $T$-dimensional feature space, optimised such that points on the same object are close to each other whereas points on different objects are far apart. Following our own preliminary experiments and recent literature~\citep{Wang2019AssociativelySI,Engelmann20CVPR,He2020LearningAM}, the dimension is set to $T\!=\!5$. The associated loss $L_{e}$ is the discriminative loss function~\citep{de2017semantic} prevalent in bottom-up instance segmentation (see Table~\ref{table:summarizesmethods}):
\begin{equation}\label{Eq:EmbeddingLossTotal}
\begin{split}
	L_{e}= &L_{e\_var}+L_{e\_dist}+0.001\cdot{L}_{e\_reg}\;,\quad\text{with}\\
	&L_{e\_var} = \frac{1}{N_{gt}}\sum_{i=1}^{N_{gt}}\frac{1}{\lvert{I_i^{gt}}\rvert}\sum_{j=1}^{\lvert{I_i^{gt}}\rvert} [||\mu_{i}-{EF}_{j}||_{1}-\delta_{v}]^{2}_+\;,\\
	&L_{e\_dist} = \frac{1}{N_{gt}(N_{gt}-1)}\mathop{\sum_{i_A=1}^{N_{gt}}\sum_{i_B=1}^{N_{gt}}}\limits_{i_A\neq{i_B}}[2\delta_{d}-||\mu_{i_A}-\mu_{i_B}||_{1}]^{2}_+\;,\\
	&L_{e\_reg} = \frac{1}{N_{gt}}\sum_{i=1}^{N_{gt}}||\mu_{i}||_1\;.
	\end{split}
\end{equation}
$N_{gt}$ denotes the total number of ground truth instances, $\lvert{I_i^{gt}}\rvert$ is the number of points in the $i$-th ground truth instance $I_i^{gt}$, $\mu_{i}$ is the mean embedding of all points in instance $I_i^{gt}$, and ${EF}_{j}$ is the predicted embedding of point $j$ in instance $I_i^{gt}$. The operators are $||\cdot||_1$ for the  $L^1$ (Manhattan) distance and $[x]_+ = \text{max}(0, x)$. The hyper-parameters are chosen as $\delta_{v}=0.5$ and $\delta_{d}=1.5$.

\noindent\textbf{Instance offset head}: Instead of a discriminative embedding, this branch directly regresses offset vectors $\mathbf{O}=\left\{O_{i}\right\}\in\mathbb{R}^{N\times{3}}$ from each 3D point's location to the centroid of the instance the point belongs to. As loss function $L_{o}$ the one proposed by  PointGroup~\citep{jiang2020pointgroup} is adopted. Besides a standard $L^1$ regression loss ($L_{o\_reg}$) between the predicted and ground truth offsets, the loss also utilises the cosine similarity ($L_{o\_dir}$) to better constrain the direction of the offset:
\begin{equation}\label{Eq:OffsetLossTotal}
\begin{split}
	L_{o}= &L_{o\_reg}+L_{o\_dir}\;,\quad\text{with}\\
   &L_{o\_reg} = \frac{1}{\sum_{i=1}^{N_{gt}}{\lvert{I_i^{gt}}\rvert}}\sum_{i=1}^{N_{gt}}\sum_{j=1}^{\lvert{I_i^{gt}}\rvert} ||O_{j}-(C_{i}-P_j)||_1\;,\\
   &L_{o\_dir} = -\frac{1}{\sum_{i=1}^{N_{gt}}{\lvert{I_i^{gt}}\rvert}}\sum_{i=1}^{N_{gt}}\sum_{j=1}^{\lvert{I_i^{gt}}\rvert} \frac{O_{j}}{||O_{j}||_2}\cdot\frac{C_{i}-P_j}{||C_{i}-P_j||_2}\;.
\end{split}
\end{equation}
Again, $N_{gt}$ is the number of ground truth instances and $\lvert{I_i^{gt}}\rvert$ is the number of points in instance $I_i^{gt}$. Furthermore, ${O}_{j}$ is the predicted offset from point $j$ to the centroid of instance $I_i^{gt}$, ${P}_{j}$ are the original 3D coordinates of point $j$, ${C}_{i}$ is the centroid of instance $I_i^{gt}$ and $||\cdot||_2$ is the $L^2$ (Euclidean) distance.

The entire neural network, including the backbone and the prediction heads, is trained from scratch by minimising a joint loss function
\begin{equation}\label{Eq:TotalLoss}
	L=L_{s}+W_{e}\cdot{L}_{e}+W_{o}\cdot{L}_{o}+W_{r}\cdot{L}_{r}\;,
\end{equation}
where $L_{r}$ is a standard $L^2$ regulariser on the network weights. $W_{e}$, $W_{o}$ and $W_{r}$ denote the weight for each loss term, respectively.

%-------------------------------------------------------------------------
\subsubsection{Post-processing}
\label{subsubSec:Post-processing}
%-------------------------------------------------------------------------

During inference, a sliding-window scheme is employed, with regularly-spaced spheres with a stride $s$, chosen such that adjacent spheres overlap and every point in the test set is processed at least once. The semantic class probabilities from overlapping spheres are averaged, and the final semantic segmentation is obtained as the point-wise $\text{argmax}$ over those average scores. Points assigned to a ``stuff'' class are removed before instance segmentation. All points assigned to ``things'' classes are embedded according to the predictions of the instance segmentation head, i.e., either by reading out their feature embedding, or by shifting them in 3D space according to the offset vectors. The embedded points are then clustered into instances. In line with other authors~\citep{zhao2021technical} our results suggest that conventional clustering as a post-process performs better than learned clustering within an end-to-end architecture. Like several others~\citep{Wang2019AssociativelySI,Lahoud20193DIS,he2021dyco3d} our study finds mean-shift to be an efficient and reliable clustering algorithm for the learned instance embedding.
On the contrary, and in line with the literature, connected components search works better to cluster points based on their shifted coordinates (offset predictions add original coordinates). To that end, one defines two thresholds $Th_{d}$ and $Th_{n}$. A set of points is considered an instance if \emph{(i)} all points have the same semantic label, \emph{(ii)} the Euclidean distances between shifted coordinates of those points are $<Th_{d}$, and \emph{(iii)} the number of points in the instance is $>Th_{n}$.

Clusters with too few points are not assigned to any instance at this stage, their instance labels will be determined later when moving back from the subsampled to the original point cloud (see below). The BlockMerging procedure of~\citet{Wang2018SGPNSG} is used to reconcile instance labels between overlapping spheres, except that a threshold $Th_{bm}$ on the IoU turned out to be a better merging criterion than the absolute number of common points. The three thresholds (hyper-parameters) are fixed, with the exception of an obvious (linear) dependency between $Th_d$ and the voxel grid size, see Table~\ref{table:parameterSettingNpm3d}.

Finally, to obtain a complete panoptic segmentation of the full, raw point cloud (i.e., undo the voxel-grid sampling), the semantic labels and instance labels are mapped back to the original points with nearest-neighbour assignment.

%%--------------------------------------------------------------------------
\subsection{Experiment details}
\label{Sec:Experiment details}
%%--------------------------------------------------------------------------

The segmentation pipeline has been implemented based on the Torch-Points3D library~\citep{Thomas2020Torch}. 4-fold cross-validation experiments were conducted on the NPM3D dataset. For each backbone, the hyper-parameters are tuned to yield the best performance, see Table~\ref{table:parameterSettingNpm3d}. The corresponding ablation studies are available in~\ref{Sec:AblationStudiesonNPM3Ddataset}.

In all experiments, the network is trained on a single NVIDIA TITAN RTX GPU for 500 epochs, with SGD with momentum as the optimiser and a base learning rate of 0.01. Each epoch comprises 3000 randomly sampled spheres, using mini-batches of 8 spheres.

\begin{table}[htbp]
\centering
\caption{Best parameter settings for the NPM3D dataset.}
\vspace{0.3cm}
\resizebox{0.5\textwidth}{!}{
\begin{tabular}{lll}
\hline
Category & Parameter & Best values / default values (unit) \\ \hline
\multirow{5}{*}{Input}
& $d$       & 0.12 (m) \\
& $R$       & 8 (m) \\
& $s$       & 8 (m) \\
& $k$       & 17500 \smallskip \\
& $K$   & \begin{tabular}[c]{@{}l@{}}3 for PointNet++\\ 4 for KPConv and sparse CNN\end{tabular} \\ \hline
\multirow{3}{*}{Training}
& $W_{e}$   & 1 \\
& $W_{o}$   & 0.1 \\
& $W_{r}$   & 0 \\ \hline
\multirow{4}{*}{Thresholds}
& $Th_d $   & $1.5\cdot{d}$ \\
& $Th_n$    & 10 \\
& $Th_{bm}$ & 0.01 \\
& Bandwidth & 0.6 \\ \hline
\end{tabular}}
\label{table:parameterSettingNpm3d}
\end{table}

%%--------------------------------------------------------------------------
\subsection{Experimental results}
\label{Sec:ExperimentalResults}
%%--------------------------------------------------------------------------

Following the pipeline described in Section~\ref{subSec:ProposedPipeline} and the experimental details in Section~\ref{Sec:Experiment details}, this research averaged the results for 4 cross-validation experiments to obtain final results for each backbone.

%--------------------------------------------------------------------------
\subsubsection{Performance of different backbones}
\label{subsec:PerformanceBasedOn3Backbones}
%--------------------------------------------------------------------------

Table~\ref{table:cross-validationresultsNpm3d} shows the quantitative results of the cross-validation experiments on NPM3D, mainly focusing on semantic segmentation mIoU and panoptic segmentation metrics. The table also records the standard deviation (std) values of two main metrics: $\text{PQ}\dag$ for panoptic segmentation and mIoU for semantic segmentation. As illustrated in Table~\ref{table:cross-validationresultsNpm3d}, KPConv outperforms the other two backbones in all metrics. Sparse CNN has the smallest standard deviations for $\text{PQ}\dag$ and mIoU, meaning that it is particularly stable across the four cross-validation folds. The PointNet++ backbone, on the other hand, shows comparatively poor performance, in particular when it comes to separating instances of ``things''. The experimental results validate the observation from Table~\ref{table:compare_backbones}, that PointNet++ ignores the geometric relationships between neighboring points (see Figure~\ref{Fig:point-basedbackbones}), resulting in poor generalization performance for large, complex scenes compared to the other two methods.

% add table for quantitative analysis
\begin{table}[]
\caption{Comparison of three backbones in terms of panoptic segmentation metrics and mIoU for semantic segmentation. For metrics $\text{PQ}\dag$ and mIoU, also the standard deviation (std) over 4 cross-validation runs is given. \textcolor{blue}{Blue} indicates which backbone performed best.}
\vspace{0.3cm}
\resizebox{\textwidth}{!}{
\begin{tabular}{l|cccc|ccc|ccc|c}
\hline
\multirow{2}{*}{Backbone} & \multicolumn{4}{c|}{panoptic segmentation} & \multicolumn{3}{c|}{\begin{tabular}[c]{@{}c@{}}panoptic segmentation\\ (``things'')\end{tabular}} & \multicolumn{3}{c|}{\begin{tabular}[c]{@{}c@{}}panoptic segmentation\\ (``stuff'')\end{tabular}} & \multicolumn{1}{c}{\begin{tabular}[c]{@{}c@{}}semantic\\segmentation\end{tabular}} \\ \cline{2-12} 
& PQ & $\text{PQ}\dag \pm \text{std}$ & RQ & SQ & PQ & RQ & SQ & PQ & RQ & SQ & \multicolumn{1}{c}{$\text{mIoU} \pm \text{std}$} \\ \hline
KPConv & \textcolor[rgb]{0,0,1}{66.4} & $\textcolor[rgb]{0,0,1}{67.0 \pm  2.3}$ & \textcolor[rgb]{0,0,1}{74.4} & \textcolor[rgb]{0,0,1}{86.7} & \textcolor[rgb]{0,0,1}{60.4} & \textcolor[rgb]{0,0,1}{65.8} & \textcolor[rgb]{0,0,1}{90.9} &\textcolor[rgb]{0,0,1}{78.4} &        \textcolor[rgb]{0,0,1}{91.6} &\textcolor[rgb]{0,0,1}{78.4} & $\textcolor[rgb]{0,0,1}{74.3 \pm 5.2}$ \\ 
Sparse CNN  &  63.0 &$65.1\pm 2.2$ & 69.9 & 84.7 & 57.9 & 63.2 & 90.4 & 73.3 & 83.3 & 73.3  & $73.2 \pm 4.0$ \\ 
PointNet++ & 40.1 & $43.3 \pm 5.4$ & 45.7 & 74.5 & 29.2 &  35.3 &  80.7 &   62.1 &   66.6 & 62.1 &   $61.9 \pm 6.9$ \\ \hline
\end{tabular}}
\label{table:cross-validationresultsNpm3d}
\end{table}

\begin{figure}[htbp]
	\centering
	\includegraphics[width=\linewidth]{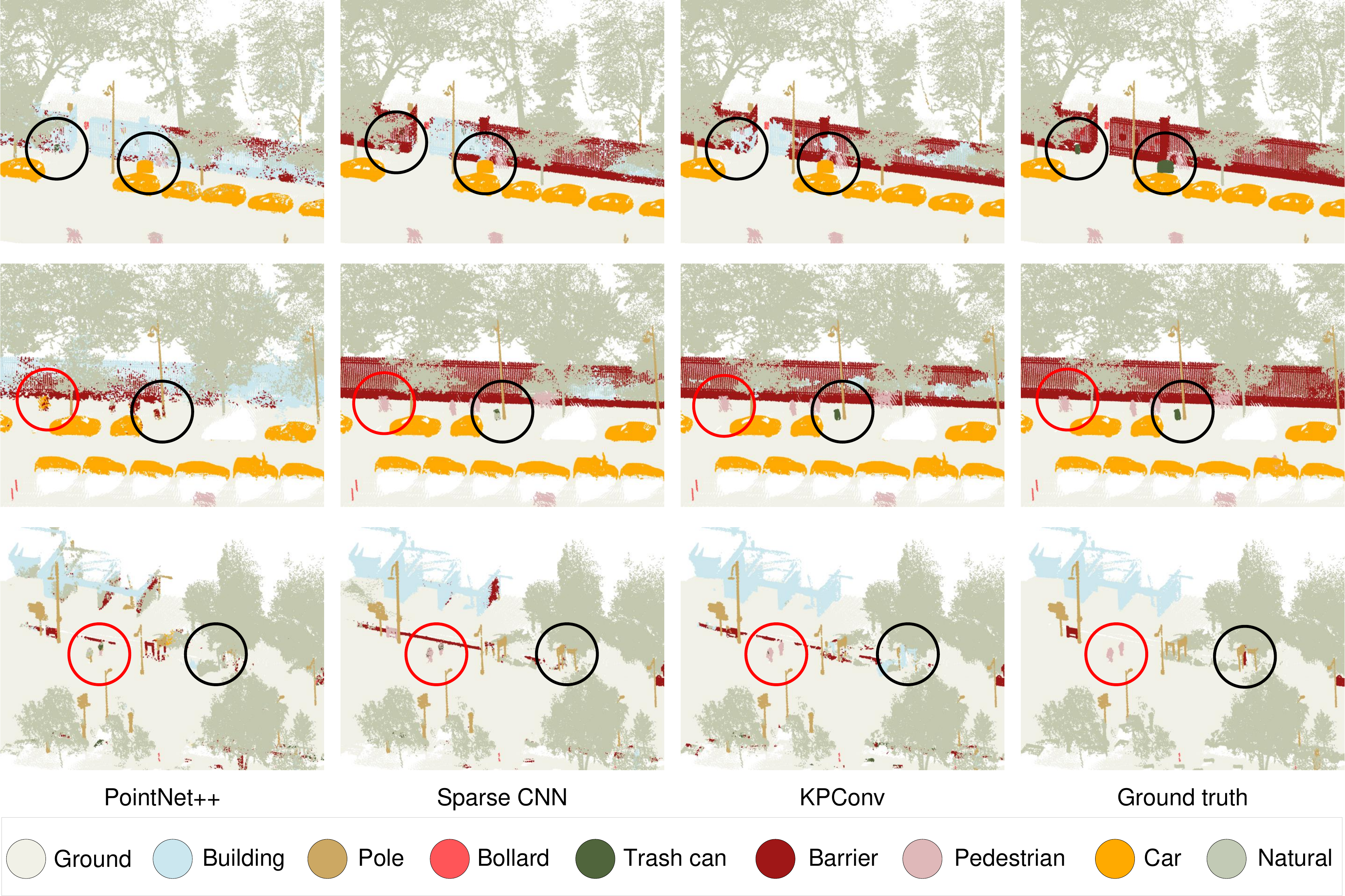}\\
	\caption{Close-up semantic segmentation results with different backbones, and ground truth.}
	\label{Fig:semanticSegmentationofCrossvalidation}
\end{figure}

\begin{figure}
  \subfigure[Semantic segmentation IoU]{
  \begin{minipage}[t]{0.47\linewidth}
    \centering
    \includegraphics[scale=0.46]{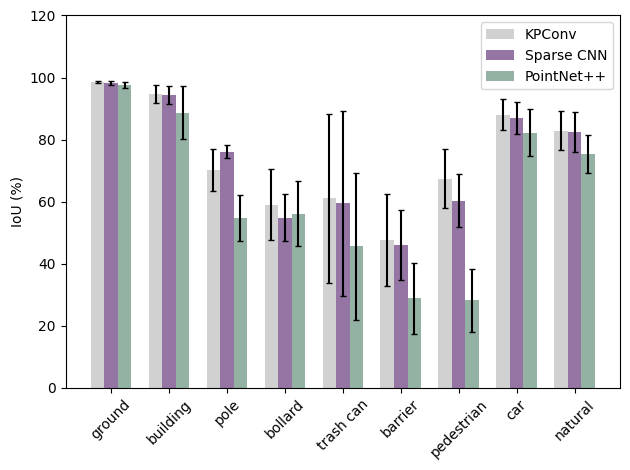}
    \vspace{-1.2cm}
    \label{Fig:crossvalidation-miouNPM3D}
  \end{minipage}%
  }
  \hspace{.15in}
  \vspace{1.5cm}
  \subfigure[Panoptic quality $\text{PQ}\dag$]{
  \begin{minipage}[t]{0.47\linewidth}
    \centering
    \includegraphics[scale=0.46]{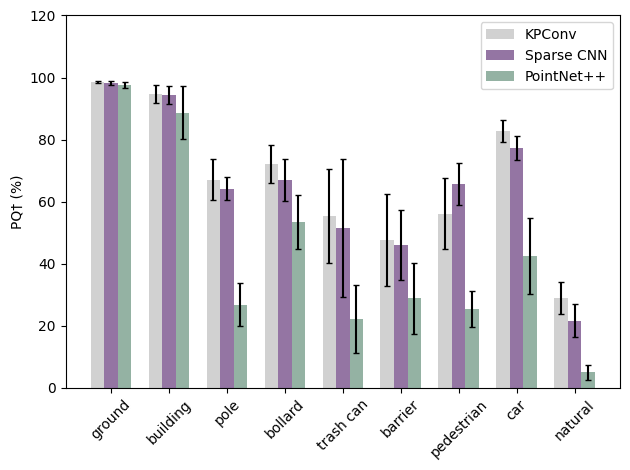}
    \vspace{-1cm}
    \label{Fig:crossvalidation-PQstarNPM3D}
  \end{minipage}%
  }
  \vspace{-2cm}
  \caption{Per-class results with 3 different backbones (mean and standard deviation of 4 cross-validation folds).}
  \label{fig:Crossvalidation-NPM3D}
\end{figure}

\begin{figure}[htbp]
	\centering
	\includegraphics[width=\linewidth]{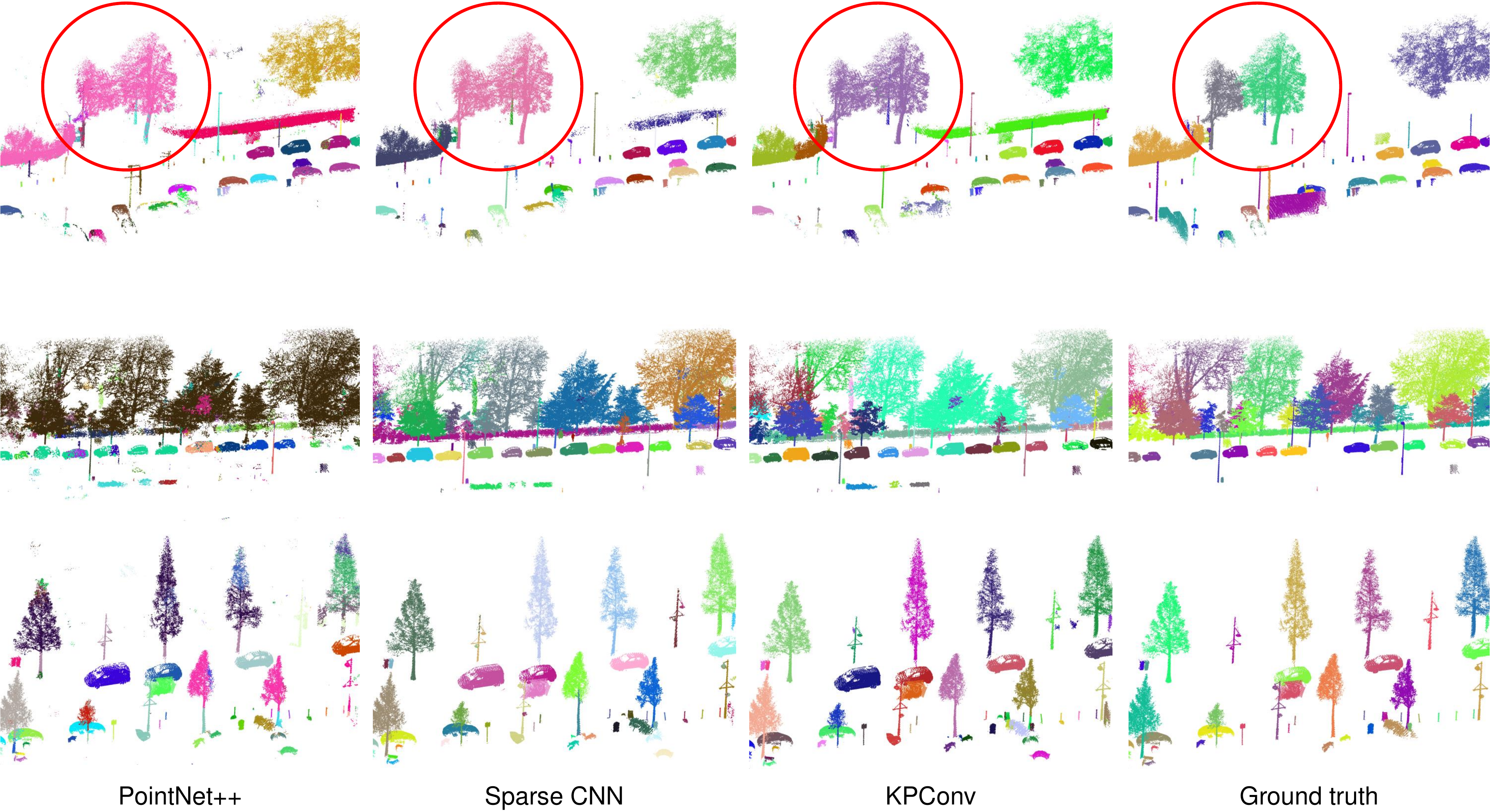}\\
	\caption{Some close-up instance segmentation qualitative results based on 3 backbones and ground truth on NPM3D dataset. Different colors represent different instance, and the colors are randomly generated.}
	\label{Fig:instanceSegmentationofCrossvalidation}
\end{figure}

Figure~\ref{Fig:crossvalidation-miouNPM3D} and Figure~\ref{Fig:crossvalidation-PQstarNPM3D} show the IoU and $\text{PQ}\dag$ as well as the standard deviation for each semantic class. 
Looking at the results, it is clear that KPConv achieves the highest IoU and $\text{PQ}\dag$ in most semantic classes. Sparse CNN reached the highest IoU for the pole class and the highest $\text{PQ}\dag$ for the pedestrian class. This can be visually confirmed by the corresponding qualitative results for semantic and instance segmentation shown in Figure~\ref{Fig:semanticSegmentationofCrossvalidation} and Figure~\ref{Fig:instanceSegmentationofCrossvalidation}. In the third row of Figure~\ref{Fig:semanticSegmentationofCrossvalidation}, the points inside the black circle marker that should belong to a pole are accurately segmented only by sparse CNN. A possible explanation is that KPConv is able to capture more complex structures, while being more sensitive to noise. For instance, in the example inside the black circle marker in the third row of Figure~\ref{Fig:semanticSegmentationofCrossvalidation}, when a street sign with a short pole is close to the tree crown, KPConv might mistake the street sign for a protruding part of the tree crown. On the other hand, sparse CNN is regularised more strictly by the native voxel resolution and therefore potentially less sensitive (Table~\ref{table:compare_backbones}), so it achieves a better segmentation in this particular case. Moreover, for moving pedestrians, the collected data often has motion artifacts (elongated ghosting or tail-like features in the point cloud). Also these can lead to over-segmentation with KPConv, due to its higher sensitivity. PointNet++ is relatively poor for all classes. In the second and third rows, marked by red circles in Figure~\ref{Fig:semanticSegmentationofCrossvalidation}, the PointNet++ backbone fails to correctly classify pedestrians, which is much worse than the other two backbones. In addition, the PointNet++ backbone often confuses barriers with buildings, see Figure~\ref{Fig:semanticSegmentationofCrossvalidation}.

The largest standard deviation was observed for the trash can class, Figure~\ref{Fig:crossvalidation-miouNPM3D}, which is due to the poor results of all three backbones for area 4. It is noted that Area 4 was collected in the city of Paris, while data for the other three areas was collected in Lille. This causes a larger domain gap (e.g., differently sized and/or shaped street furniture), and consequently poorer semantic segmentation in Area 4. As shown by the black circle markers in the first row of Figure~\ref{Fig:semanticSegmentationofCrossvalidation}, none of the three backbones can correctly classify the points belonging to trash cans. The  black circle marker in the second line of Figure~\ref{Fig:semanticSegmentationofCrossvalidation} shows a case that can only be segmented well by KPConv. The natural class gets satisfactory IoU but very low $\text{PQ}\dag$ values, for all backbones. This is because the vegetation instances are poorly segmented, especially when trees are close to each other and overlap, see first and second lines of Figure~\ref{Fig:instanceSegmentationofCrossvalidation}.

% visualization results

% add figure for qualitative analysis

%--------------------------------------------------------------------------
%\subsubsection{Compare 2 kinds of features for instance clustering}
\subsubsection{Comparison between offset and embedding-based clustering}
\label{subsec:2featuresForInstanceClustering}
%--------------------------------------------------------------------------

Table~\ref{table:instanceseg2features-NPM3D} shows a comparison between instance (and panoptic) segmentation obtained with either the offset or the embedding head. In general, instance segmentation results based on embedding features are significantly better than those based on shifted coordinates for the KPConv and sparse CNN backbones. In the case of the PointNet++, performance based on embedding features is only slightly better than the one with shifted coordinates ($2.4\%$ higher $\text{PQ}\dag$).

\begin{table}[]
\caption{To compare 2 different kinds of features for instance clustering based on 3 backbones on NPM3D dataset by using panoptic segmentation metrics and instance segmentation metrics. \textcolor{blue}{Blue} indicates which instance strategy (embedding or offset prediction) performed better, \textbf{bold} font marks the best result per column.}
\vspace{0.3cm}
\resizebox{\textwidth}{!}{
\begin{tabular}{ll|ccccc|cccc|ccc}
\hline
\multirow{2}{*}{backbone} & \multirow{2}{*}{instance} & \multicolumn{5}{c|}{instance segmentation} & \multicolumn{4}{c|}{panoptic segmentation} & \multicolumn{3}{c}{\begin{tabular}[c]{@{}c@{}}panoptic segmentation\\ (``things'')\end{tabular}} \\ \cline{3-14}
& & mCov  & mWCov  & mPrec  & mRec & F1 & PQ & PQ\dag & RQ & SQ & PQ & RQ & SQ  \\ \hline
\multirow{2}{*}{KPConv} & offset  & 70.5 & 73.2 & 48.8 & 72.6 & 58.2 & 60.0 & 60.7 & 67.7 & 85.9 & 50.9 & 55.8 &89.7 \\  
& embed & \textcolor[rgb]{0,0,1}{72.8} & \textcolor[rgb]{0,0,1}{76.1} & \textcolor[rgb]{0,0,1}{\bf 60.3} & \textcolor[rgb]{0,0,1}{76.2} & \textcolor[rgb]{0,0,1}{\bf 67.2} & \textcolor[rgb]{0,0,1}{\bf 66.4} & \textcolor[rgb]{0,0,1}{\bf 67.0} &\textcolor[rgb]{0,0,1}{\bf 74.4} & \textcolor[rgb]{0,0,1}{\bf 86.7} & \textcolor[rgb]{0,0,1}{\bf 60.4} & \textcolor[rgb]{0,0,1}{\bf 65.8} & \textcolor[rgb]{0,0,1}{\bf 90.9} \\ \cline{3-14}
\multicolumn{1}{c}{\multirow{2}{*}{Sparse CNN}} & offset & 70.1 & 73.5 & 52.0 & 72.9 & 60.5 & 59.7 & 61.8 & 66.7 & 84.0 & 52.9 & 58.3 & 89.4 \\  
\multicolumn{1}{c}{} & embed & \textcolor[rgb]{0,0,1}{\bf 73.5} & \textcolor[rgb]{0,0,1}{\bf 77.3} &\textcolor[rgb]{0,0,1}{55.7} &
\textcolor[rgb]{0,0,1}{\bf 77.3} &\textcolor[rgb]{0,0,1}{64.6} & \textcolor[rgb]{0,0,1}{63.0} & \textcolor[rgb]{0,0,1}{65.1} &
\textcolor[rgb]{0,0,1}{69.9} & \textcolor[rgb]{0,0,1}{84.7} &
\textcolor[rgb]{0,0,1}{57.9} & \textcolor[rgb]{0,0,1}{63.2} &
\textcolor[rgb]{0,0,1}{90.4} \\ \cline{3-14}
\multicolumn{1}{c}{\multirow{2}{*}{PointNet++}} & offset                   &\textcolor[rgb]{0,0,1}{54.9} & \textcolor[rgb]{0,0,1}{53.9} & 27.5        &\textcolor[rgb]{0,0,1}{58.0} & 36.3 &  37.7 & 40.9 & 42.6 & \textcolor[rgb]{0,0,1}{75.5} &  25.6 & 30.6 & \textcolor[rgb]{0,0,1}{82.2}  \\
\multicolumn{1}{c}{} & embed & 47.8 & 51.7 &\textcolor[rgb]{0,0,1}{30.8} &47.5 & \textcolor[rgb]{0,0,1}{37.0} &  \textcolor[rgb]{0,0,1}{40.1} &
\textcolor[rgb]{0,0,1}{43.3} & \textcolor[rgb]{0,0,1}{45.7} &  74.5 & \textcolor[rgb]{0,0,1}{29.2} &  \textcolor[rgb]{0,0,1}{35.3} &  80.7 \\ \hline
\end{tabular}}
\label{table:instanceseg2features-NPM3D}
\end{table}

\begin{figure}
  \subfigure[]{
  \begin{minipage}[t]{0.47\linewidth}
    \centering
    \includegraphics[scale=0.46]{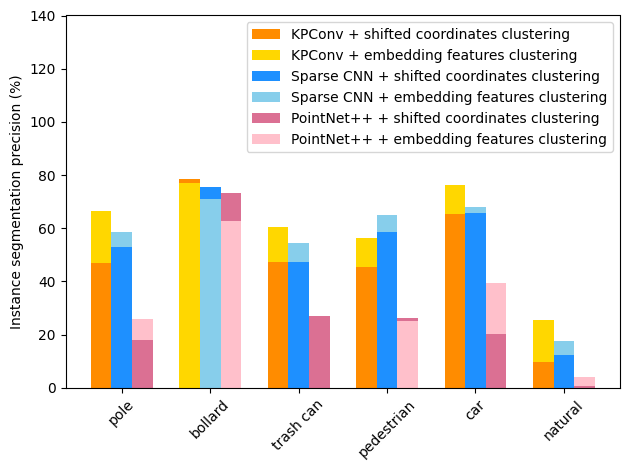}
    \vspace{-1.2cm}
    \label{Fig:precInstanceNPM3D}
  \end{minipage}%
  }
  \hspace{.15in}
  \vspace{1.5cm}
  \subfigure[]{
  \begin{minipage}[t]{0.47\linewidth}
    \centering
    \includegraphics[scale=0.46]{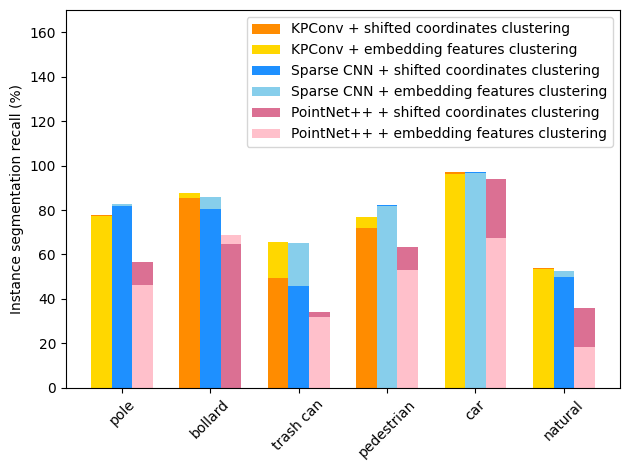}
    \vspace{-1cm}
    \label{Fig:recInstanceNPM3D}
  \end{minipage}%
  }
  \vspace{-2cm}
  \caption{Comparison of instance segmentation precision and recall for each ``things'' class, based on 3 backbones and 2 types of clustering features.}
  \label{fig:instancefeatures-NPM3D}
\end{figure}

\begin{figure}[htbp]
	\centering
	\includegraphics[width=\linewidth]{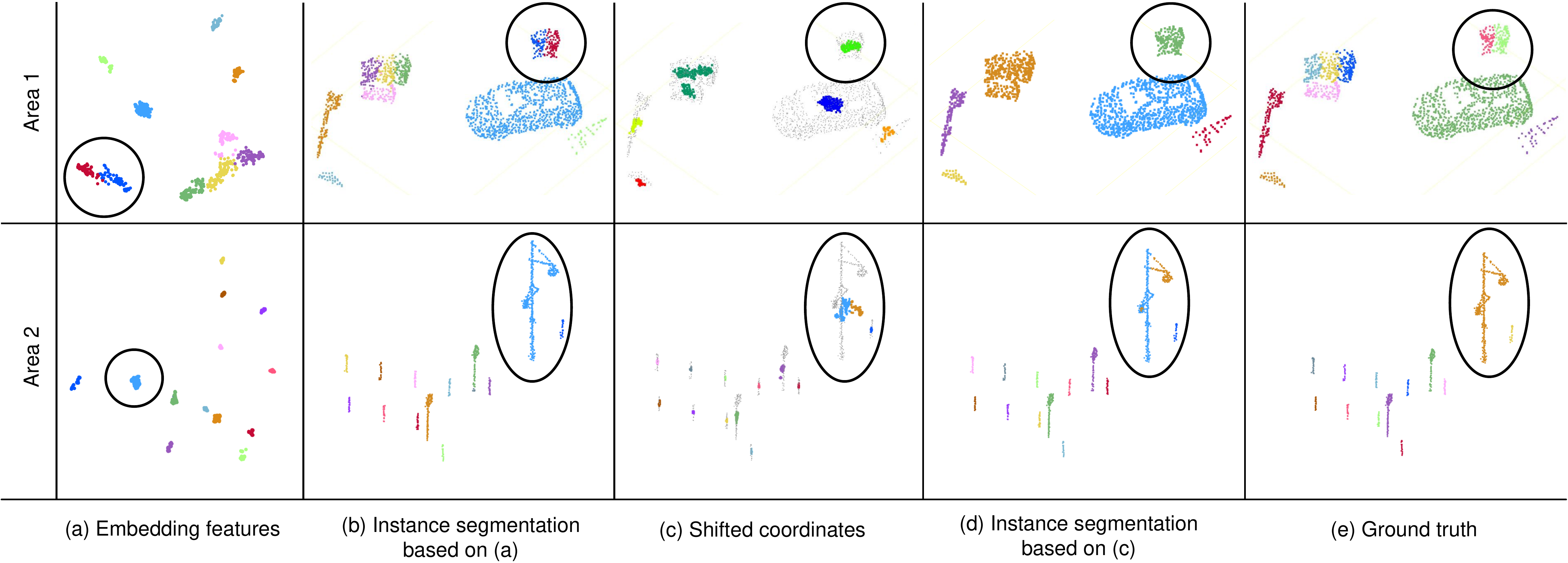}\\
    	\caption{Take 2 close-ups of instance segmentation based on KPConv backbone as examples. Column (a) displays the embedding features in 2D space by PCA algorithm, where the colors are randomly generated and the points of different instances have different colors. (b) Instance segmentation based on the clustering of embedding features, where the colors of points keep correspondence with column (a). (c) The gray points are the original points, and the colored points indicate the shifted coordinates, where the colors are randomly generated and the points of different instances have different colors. (d) Instance segmentation based on shifted coordinates clustering. (e) shows the ground truth of instance segmentation, where the colors are randomly generated and the points of different instances have different colors.}
	\label{Fig:compare2DifferentFeatures}
\end{figure}

\begin{figure}[htbp]
	\centering
	\includegraphics[width=\linewidth]{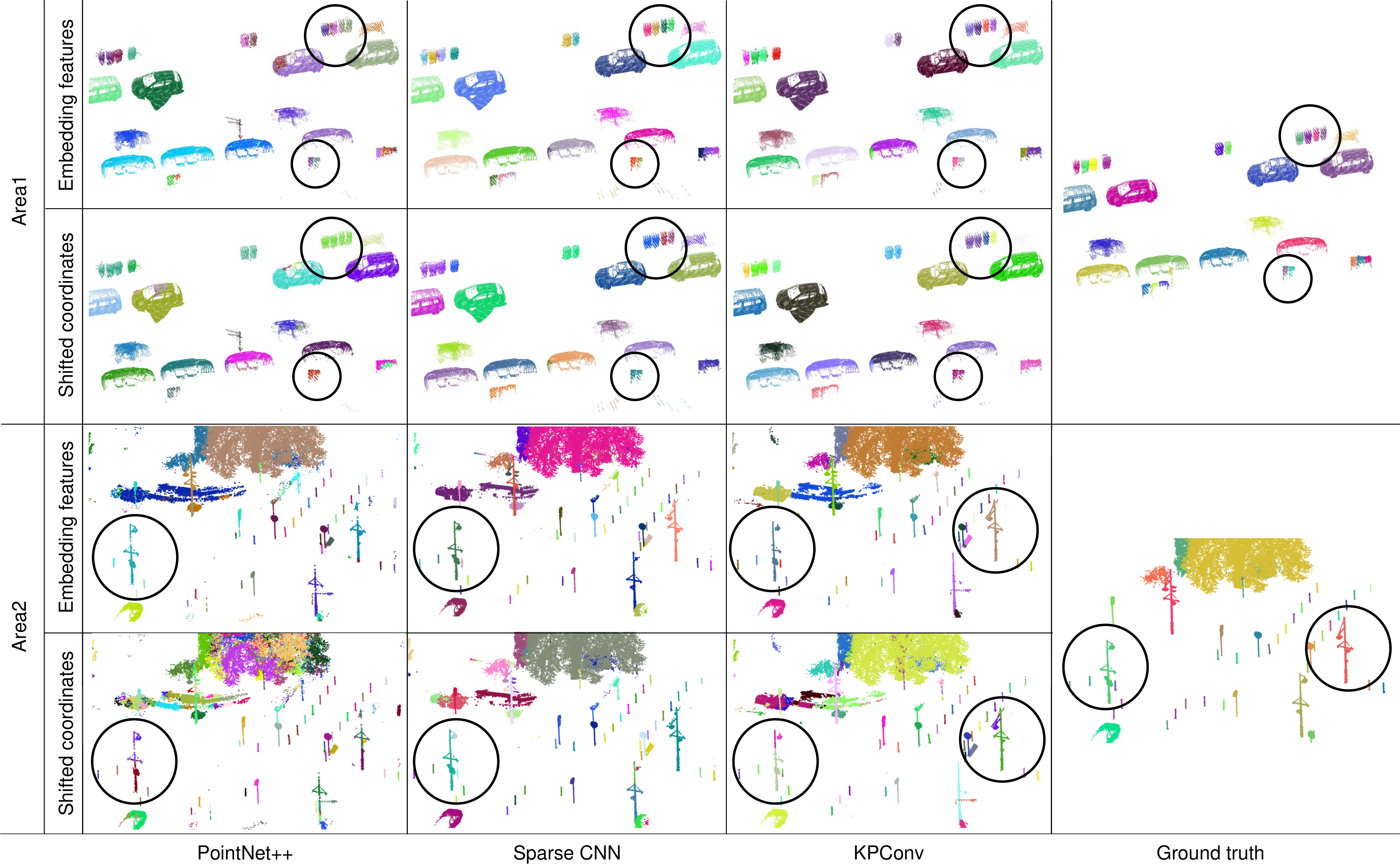}\\
	\caption{Qualitative results of instance segmentation based on 2 different features and 3 backbone networks. Different colors indicate different instances, and the colors are randomly generated.}
	\label{Fig:compare2DifferentFeatures2}
\end{figure}

Figure~\ref{Fig:precInstanceNPM3D} and Figure~\ref{Fig:recInstanceNPM3D} show the precision and recall per backbone and instantiation head. Clustering based on embedding features generally achieves better instance segmentation precision than shifted coordinates, except for the bollard class. For the KPConv and sparse CNN backbones, instance segmentation recall based on embedding features is also higher or approximately equal to the one based on shifted coordinates. For the PointNet++ backbone its the opposite, here the recall is higher if shifted coordinates are used. 
Figure~\ref{Fig:compare2DifferentFeatures} compares both clustering schemes on two example scenes based on the KPConv backbone. 
In the first row (Area 1), black circle marker points out two trash cans that are very close to each other. Here the learned embedding features can better distinguish them, while the shifted coordinates are not able to separate them (under-segmentation).
The second row (Area 2) shows a streetlight inside the black circle marker. It can be seen that the embedding features belonging to this instance are very compact, and thus the instance can be easily and completely segmented from other instances. However, clustering based on shifted coordinates splits the lamp head and the lamp pole into two instances (over-segmentation). For more visual examples see Figure~\ref{Fig:compare2DifferentFeatures2}. 

%--------------------------------------------------------------------------
\subsubsection{Runtime analysis}
\label{subsec:RuntimeAnalysis}
%--------------------------------------------------------------------------

Runtime was measured for training and testing to compare the computational complexity of the different backbones. Table~\ref{table:trainingtimeNpm3d} shows the time required to load the data and to perform the actual computation for one training epoch. The computation includes the following operations: forward pass through the complete network, calculation of losses and gradients, and update of network weights. Data loading involves unpacking the input data from the data loader and performing the necessary preprocessing steps. As shown in Table~\ref{table:trainingtimeNpm3d}, the KPConv backbone has a longer data loading time because, in addition to the data augmentation, it requires time to down-sample the point cloud and precompute neighborhoods on the CPU to speed up the forward pass.

\begin{table}[htbp]
\caption{\added[id=R2]{Training time of main stages for three different backbones. All times were measured on the same hardware (8-core CPU with 4GB of memory per processor core and a single Nvidia Titan RTX GPU). The table shows the average duration of one epoch.}}
\vspace{0.3cm}
\centering
%\resizebox{\textwidth}{!}{
\begin{tabular}{|l|cc|c|}
\cline{1-4}
\multicolumn{1}{|c|}{\multirow{2}{*}{Backbone}} & \multicolumn{3}{c|}{training time (sec / epoch)}\\
\cline{2-4} & computation               & data loading              &   total \\ \hline
KPConv                    & 99.9              & 755.8               & 855.6 \\
Sparse CNN                & 84.6              & 173.4                & 258.0 \\
PointNet++                & 83.8              & 142.7                & 226.5 \\ \hline
\end{tabular}
%}
\label{table:trainingtimeNpm3d}
\end{table}

Table~\ref{table:inferencetimeNpm3d} shows the runtime for different inference stages. The time for the forward pass depends on the backbone network, where PointNet++ takes the shortest time and Sparse CNN takes the longest. Instance clustering and block merging are independent of the backbone, so the time for these two steps given in Table~\ref{table:inferencetimeNpm3d} is the average over the 4-fold cross-validation experiments for all three backbones. Clustering time is measured for embedding features and shifted coordinates. Clustering of embedding features is done with the mean-shift implementation of scikit-learn~\citep{scikit-learn} which runs on CPU. Connected-components clustering of shifted coordinates employs the Torch-Points3D~\citep{Thomas2020Torch} implementation, which uses the algorithm proposed in PointGroup~\citep{jiang2020pointgroup}. Neighborhood search is done on the GPU, while cluster assignment is done on the CPU. Overall the time required for clustering based on feature embedding is almost five times longer than for shifted coordinates.

\begin{table}[htbp]
\caption{\added[id=R2]{Inference time of main stages for three different backbones on NPM3D dataset. All times were measured with the same hardware (8-core CPU with 4GB of memory per processor core and a single Nvidia Titan RTX GPU).}}
\vspace{0.3cm}
%\resizebox{\textwidth}{!}{
\begin{tabular}{|l|c|cc|c|cc|}
\cline{1-7}
\multicolumn{1}{|c|}{\multirow{4}{*}{Backbone}}
& \multicolumn{6}{c|}{inference times (sec / million points)} \\
\cline{2-7}
& backbone fwd pass &
\multicolumn{2}{c|}{instance clustering} &
\multicolumn{1}{c|}{block merging} &
\multicolumn{2}{c|}{total} \\
& (GPU) & \multicolumn{2}{c|}{(GPU/CPU)} & (CPU) & & \\
& & embed & offset & & embed & offset \\
\hline
KPConv
& 5.0
& \added[id=R2]{39.1}
& \added[id=R2]{8.2} & \added[id=R2]{1.2} & 45.3 & 14.4 \\
Sparse CNN & 9.8 & 39.1
& 8.2 & 1.2 & 50.1 & 19.2 \\
PointNet++ & 3.2 & \added[id=R2]{39.1}
& \added[id=R2]{8.2} & \added[id=R2]{1.2} & 43.5 & 12.6 \\ \hline
\end{tabular}
%}
\label{table:inferencetimeNpm3d}
\end{table}

%-------------------------------------------------------------------------
\section{Conclusions}
\label{Sec:Conclusions}

%-------------------------------------------------------------------------

A comprehensive review has been presented for the task of 3D panoptic segmentation, including suitable 3D backbone networks, instance segmentation strategies, existing datasets and evaluation metrics.
So far, a systematic comparison of panoptic segmentation methods for outdoor mobile mapping point clouds has been lacking. Therefore, instances have been annotated in the NPM3D dataset, and a modular panoptic segmentation pipeline has been implemented for such mobile mapping point clouds, with a choice of three different backbones representative of different schools of point cloud processing, and two representative instance segmentation strategies. In a comprehensive experimental evaluation, it was found that the KPConv backbone achieved the highest panoptic segmentation performance, but also has the highest runtime. PointNet++ has the lowest computational cost for training and inference, but yields relatively poor results compared to the others. The sparse voxel CNN offers a good trade-off between computation time and segmentation performance.
Moreover, our experiments showed that for our mobile mapping dataset instance segmentation by clustering embedding features gives better results than clustering with shifted coordinates,
regardless of the backbone used. 

%  ADD ACKNOWLEDGMENTS LATER FOR FINAL PAPER AFTER ACCEPTANCE
%==============================================
%\section*{Acknowledgment}
%==============================================
%This study was partly funded by the Chinese Scholarship Council (CSC, No.201906270261).

\small
\bibliographystyle{elsarticle-harv}
\bibliography{isprsbib-InstanceSeg}

\clearpage
\appendix
\section{}
% the \\ insures the section title is centered below the phrase: AppendixA

%--------------------------------------------------------------------------
\subsection{Ablation studies on NPM3D}
\label{Sec:AblationStudiesonNPM3Ddataset}

Ablation studies were performed on NPM3D to analyze the influence of hyper-parameters on the results. These include ablation of the regularization loss, the size of input samples and the type of input features. The description of the parameters can be found in Table~\ref{table:allParametersNotationAndDescription} in the main paper. In each experiment, the values of parameters that are not specifically mentioned are set to the default values in Table~\ref{table:parameterSettingNpm3d}.

\begin{figure}
  \subfigure[$\text{PQ}\dag$ of PointNet++, sparse CNN and KPConv with and without weight regularization.]{
  \begin{minipage}[t]{0.47\linewidth}
    \centering
    \includegraphics[scale=0.46]{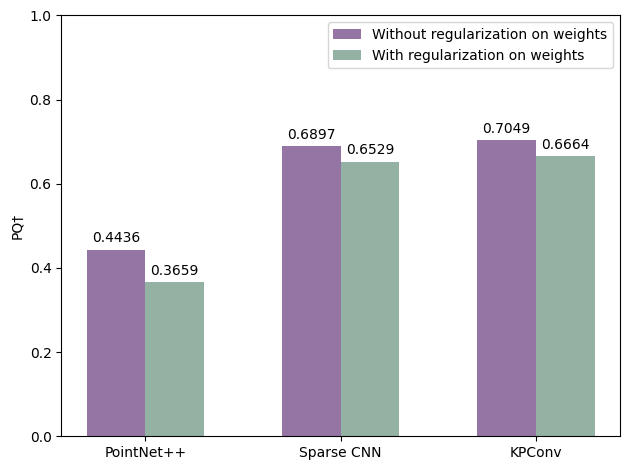}
    \vspace{-1.2cm}
    \label{Fig:regularizationNPM3D}
  \end{minipage}%
  }
  \hspace{.15in}
  \subfigure[Panoptic segmentation performance with 5 different voxel sizes (sparse CNN backbone).]{
  \begin{minipage}[t]{0.47\linewidth}
    \centering
    \includegraphics[scale=0.46]{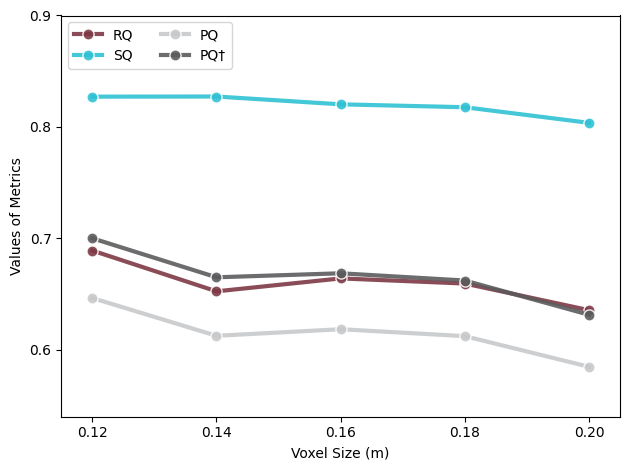}
    \vspace{-1.2cm}
    \label{Fig:voxelsizeNPM3D}
  \end{minipage}
  }
  \quad
  \vspace{1.5cm}
  \subfigure[Panoptic segmentation performance when changing the radius of the input neighborhood (sparse CNN backbone).]{
  \begin{minipage}[t]{0.47\linewidth}
    \centering
    \includegraphics[scale=0.46]{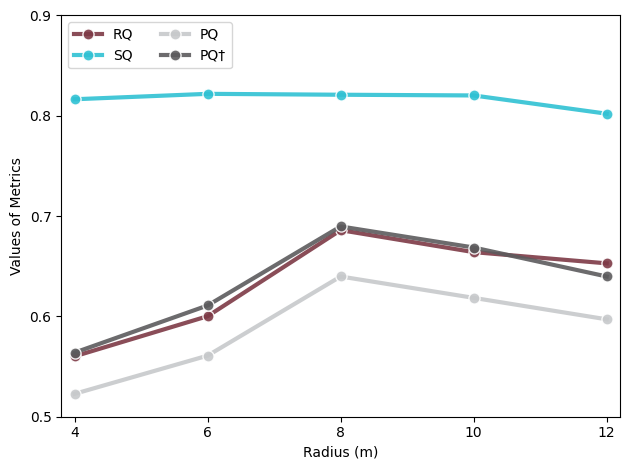}
    \vspace{-1cm}
    \label{Fig:radiusNPM3D}
  \end{minipage}%
  }
  \hspace{.15in}
  \vspace{1.5cm}
  \subfigure[$\text{PQ}\dag$ of PointNet++, sparse CNN and KPConv with or without absolute $Z$ coordinates as additional input.]{
  \begin{minipage}[t]{0.47\linewidth}
    \centering
    \includegraphics[scale=0.46]{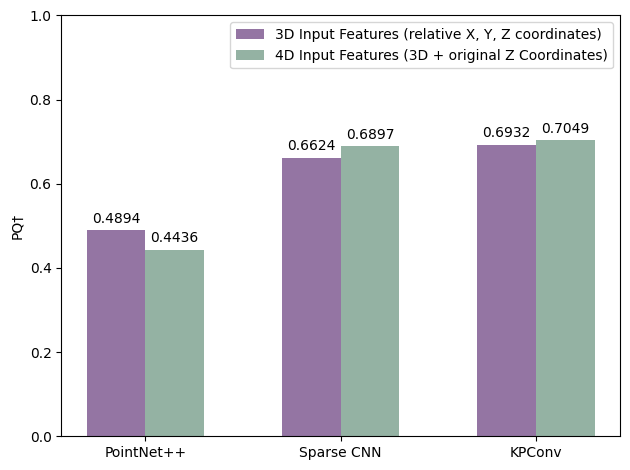}
    \vspace{-1cm}
    \label{Fig:inputfeaturesNPM3D}
  \end{minipage}
  }
  \vspace{-3.2cm}
  \caption{Ablation studies on NPM3D dataset}
  \label{fig:AblationNPM3D}
\end{figure}

\noindent
\textbf{Regularisation loss}: To investigate how weight regularisation affects the panoptic segmentation performance, the regularisation term $L_r$ is added with a weight of $W_r=0.001$ (Equation~\ref{Eq:TotalLoss}) to the total cost function for three backbone networks. For a fair comparison, the input size and input features are the same for the experiments with and without regularization loss term: $d$=0.16m, $R$=8m, $K$=4, see Table~\ref{table:allParametersNotationAndDescription}, and use $\text{PQ}\dag$ as the main quantitative indicator. As shown in Figure~\ref{Fig:regularizationNPM3D}, a higher PQ can be obtained without regularization loss term for all 3 backbones.

\noindent
\textbf{Size of input samples}: Based on the sparse CNN backbone, the ablation study explores the effect of different voxel size $d$ and sphere radius $R$ on panoptic segmentation performance. For all experiments $K=4$ and $W_r=0$. For Figure~\ref{Fig:voxelsizeNPM3D}, $R$ is fixed to $10\,$m and $d$ is varied between 0.12$\,$m and 0.2$\,$m. Smaller $d$ (i.e., higher point density) yields better performance. For Figure~\ref{Fig:radiusNPM3D} $d$ is fixed to $0.16\,$m and the sphere radius varies between 4$\,$m and 12$\,$m. The best performance is reached with $R=8\,$m. Apparently, a too large context radius makes the learning harder.

\noindent
\textbf{Input features}: For the NPM3D dataset, the performance of the three different networks was compared depending on the available input features. One trial used only the relative $(X,Y,Z)$ coordinates ($K=3$), and one used the relative $(X,Y,Z)$ and original $Z$ coordinates ($K=4$). Since RGB information is not available, it was not possible to conduct experiments for $K=6$ and $K=7$. All experiments use a uniform setting: $R=8$\,m, $W_r=0$ and $d=0.16\,$m, see Table~\ref{table:allParametersNotationAndDescription}.
The results are shown in Figure~\ref{Fig:inputfeaturesNPM3D}. For the PointNet++ backbone, the variant without the original $Z$ coordinates ($K$=3) achieves higher $\text{PQ}\dag$, while for KPConv and sparse CNN, 
adding the original $Z$-coordinates ($K$=4) brings a slight improvement.

%--------------------------------------------------------------------------
\subsection{Experiments on S3DIS dataset}
\label{Sec:ExperimentsonS3DISdataset}
%--------------------------------------------------------------------------

So far, panoptic segmentation has mostly been tested on indoor point clouds. That setting has quite different characteristics from mobile street mapping and is not the focus of our work, nevertheless the pipeline is also tested on S3DIS, so as to better place it in context.
Following a large part of the literature, Area 5 of the dataset serves as test set for the experiments.
As point density, object size etc.\ are very different from the outdoor setting, the hyper-parameters are optimised with a set of ablation studies like the ones above. Note that $\text{PQ}\dag$ is equal to PQ since the S3DIS dataset contains instance labels for all semantic categories. Therefore, we only show the PQ score as the main evaluation measure.

\begin{figure}
  \subfigure[$\text{PQ}$ acquired by PointNet++, sparse CNN and KPConv backbones with regularization loss term or without regularization loss term.]{
  \begin{minipage}[t]{0.5\linewidth}
    \centering
    \includegraphics[scale=0.5]{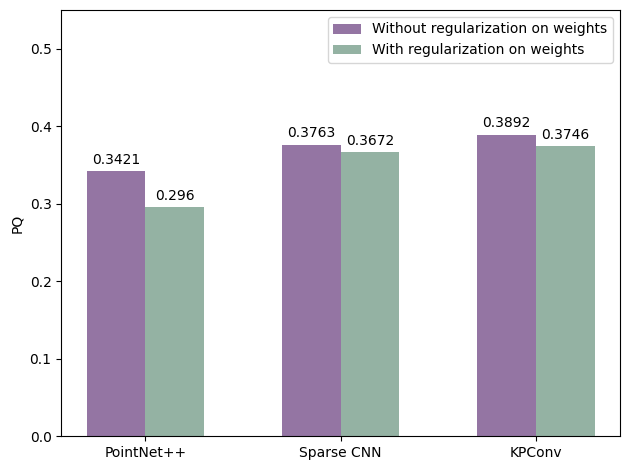}
    \vspace{-1.2cm}
    \label{Fig:regularizationS3DIS}
  \end{minipage}%
  }
  \hspace{.15in}
  \subfigure[Based on sparse CNN backbone network, the variation of 4 panoptic segmentation metrics ($\text{RQ}$, $\text{SQ}$ and $\text{PQ}$) with 5 different voxel sizes for sub-sampling of the input point cloud.]{
  \begin{minipage}[t]{0.5\linewidth}
    \centering
    \includegraphics[scale=0.5]{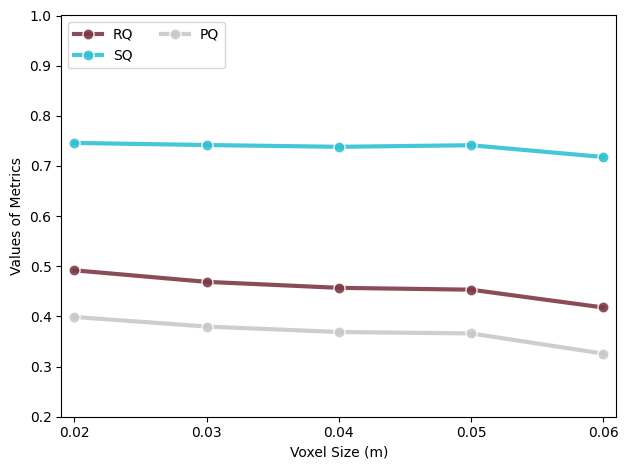}
    \vspace{-1.2cm}
    \label{Fig:voxelsizeS3DIS}
  \end{minipage}
  }
  \quad
  \vspace{1.5cm}
  \subfigure[Based on sparse CNN backbone network, the variation of 4 panoptic segmentation metrics ($\text{RQ}$, $\text{SQ}$ and $\text{PQ}$) with 5 different input radius of spherical area.]{
  \begin{minipage}[t]{0.5\linewidth}
    \centering
    \includegraphics[scale=0.5]{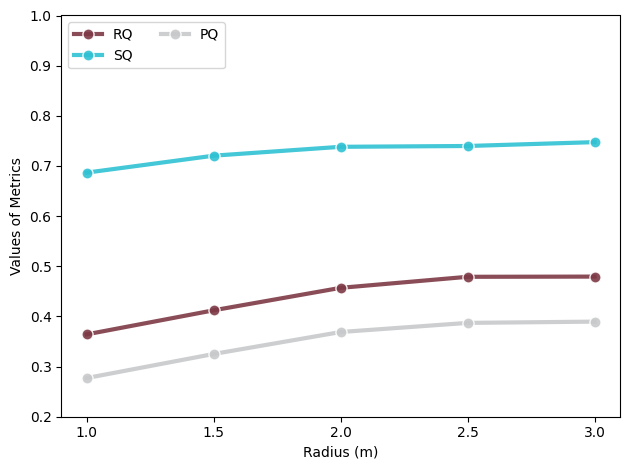}
    \vspace{-1cm}
    \label{Fig:radiusS3DIS}
  \end{minipage}%
  }
  \hspace{.15in}
  \vspace{1.5cm}
  \subfigure[$\text{PQ}$ acquired by PointNet++, sparse CNN and KPConv backbones with different input features.]{
  \begin{minipage}[t]{0.5\linewidth}
    \centering
    \includegraphics[scale=0.5]{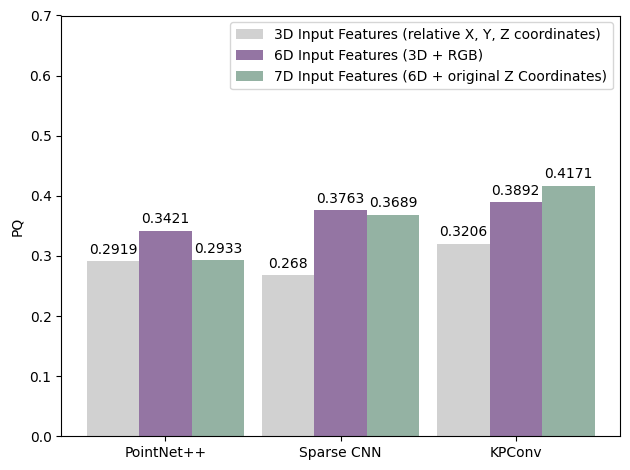}
    \vspace{-1cm}
    \label{Fig:inputfeaturesS3DIS}
  \end{minipage}
  }
  \vspace{-3.2cm}
  \caption{Ablation studies on S3DIS dataset}
  \label{fig:AblationS3DIS}
\end{figure}

\noindent\textbf{Ablation on regularization loss}: To compare the influence of the regularization loss term, the pipeline is run with radius $R$=2m and the voxel size $d$=0.04 for all backbones. Input features for all three backbones are relative $(X,Y,Z)$ and RGB ($K$=6). For the experiments with regularization loss term, $W_r=0.001$ for all backbones. As shown in Figure~\ref{Fig:regularizationS3DIS}, dropping the regularization yields higher PQ scores.

\noindent\textbf{Ablation on the sizes of input samples}: The effect of input sphere size $R$ and voxel size $d$ is assessed with the help of three panoptic segmentation metrics. The first experiment, shown in Figure~\ref{Fig:voxelsizeS3DIS}, uses a fixed sphere radius $R$=2m and examines the voxel size with five different values of $d$, ranging from 0.02m to 0.06m (with no regularization). The results show that the smaller $d$, the higher the values for all metrics of panoptic segmentation, which is consistent with the results of the NPM3D dataset (Figure~\ref{Fig:voxelsizeNPM3D}). 
In the next experiment, the voxel size is fixed to $d$=0.04m and five different $R$ values are tested, ranging from 1m to 3m. Here, the performance was improved with increasing $R$. For the S3DIS dataset, the GPU memory restricts the range of $R$ and $d$ values.

\noindent\textbf{Ablation on the input features}: In this experiment only the input features are changed, with fixed $R$=2m, $d$=0.04m, $W_r$=0. All backbones are trained and tested with three different feature sets: (1) relative $(X,Y,Z)$ ($K$=3), (2) relative $(X,Y,Z)$ and RGB ($K$=6), (3) relative $(X,Y,Z)$ and original $Z$ and RGB ($k$=7). As shown in Figure~\ref{Fig:inputfeaturesS3DIS}, good results are already obtained for all three backbones when only the relative $(X,Y,Z)$ coordinates ($K$=3) are used as input. Furthermore, all methods improve the PQ when the RGB information is available in addition to the relative point coordinates ($K$=6). Finally, adding the original $Z$ coordinate ($K$=7) improves the performance only for the KPConv backbone, all other methods perform worse when this feature is presented.

%\added{Based on the previous experiments, Table~\ref{table:parameterSettingS3DIS} shows the best hyperparameters for the different backbones.}
The best-performing parameters are shown in Table~\ref{table:parameterSettingS3DIS}. The results of our pipeline, again with 3 different backbones and 2 instance clustering methods, are shown in Table~\ref{table:S3DISArea5Compare}.  Note that in S3DIS there are no ``stuff'' classes, so PQ and PQ$\dag$ are identical.
Overall, the results show that our experimental pipeline is competitive with the state of the art also for indoor scenes. Note, though, that, for several reasons, the quantitative results from the literature are not directly comparable with ours.
\begin{enumerate}
    \item Many authors concentrate exclusively on instance segmentation and tune for that purpose, to the point that they do not even show quantitative results for semantic segmentation or panoptic segmentation.
    \item Previous work aimed only at indoor scenes~\citep{Engelmann20CVPR,jiang2020pointgroup,he2021dyco3d,chen2021hierarchical} assumes that the separation into rooms is given a-priori and feed individual rooms to the network, thus providing the best possible amount of context. Whereas our pipeline, designed with outdoor scenes in mind that lack a natural separation into self-contained sub-regions, has no prior knowledge of room boundaries and uses generic, spherical neighbourhoods.
    \item Many recent methods rely on additional post-processing steps to improve instance segmentations. E.g., DyCo3D~\citep{he2021dyco3d} and HAIS~\citep{chen2021hierarchical} predict binary masks for each cluster candidate to refine the cluster boundaries. SoftGroup~\citep{Vu2022SoftGroup} allows 3D points to be assigned to multiple semantic classes, thus mitigating instance segmentation errors due incorrect semantic labels. And PointGroup~\citep{jiang2020pointgroup} trains an additional network (called ScoreNet) on top of the actual segmentation pipeline, whose only purpose is to score all candidate clusters, such that it can generate many redundant candidate clusters and perform non-maximum suppression based on the scores. All those post-processing methods are orthogonal to the actual panoptic segmentation pipeline.
\end{enumerate}
As a baseline for our scenario, suitably adapted versions of PointGroup were trained, using its existing implementation in Torch-Points3D.
First, our PointGroup variant does not use the a-priori segmentation into rooms, and instead classifies and merges spherical neighbourhoods. Note, to fit into GPU memory (of the same GPU also used in all our other experiments) the voxel size had to be set to 3$\,$cm (rather than 2$\,$cm in the original paper).
Second, the original PointGroup method allows points to be assigned to multiple instances, or to no instance at all. To turn it into a true panoptic segmentation problem, our version ensures that each point is assigned to exactly one instance and one semantic class. Points without instance label are merged into the instance of their nearest properly labeled neighbour.
Points with multiple labels are handled as in the original PointGroup method, with a ScoreNet to assess the instance candidates, then keep the instance with the highest score. %
Additionally, a version without ScoreNet post-processing is also tested (denoted by \sout{ScoreNet}), where ambiguous points are instead assigned to the instance consistent with their most probable semantic class label.

In summary, also on S3DIS our method is competitive with PointGroup, a state-of-the-art instance segmentation method, in terms of panoptic and semantic segmentation performance. The semantic segmentation results are almost the same (except for the weaker PointNet++ backbone). In terms of instance segmentation, PointGroup without the additional ScoreNet is significantly worse (\textgreater 8 percent points difference). With ScoreNet, but without the specialisation to individual rooms, the performance is practically the same as for our pipeline.
The table also shows the result given in the original PointGroup paper, and several other recent instance segmentation works. When tuned to indoor environments the full PointGroup does have the upper hand. Also 3D-MPA~\citep{Engelmann20CVPR} reaches a similar performance, confirming the difference between the indoor room and outdoor street scenarios.

\begin{table}[htbp]
\centering
\caption{Best parameter settings for S3DIS dataset.}
\vspace{0.3cm}
%\resizebox{0.5\textwidth}{!}{
\begin{tabular}{lll}
\hline
Category & Parameter & Best values / default values (unit) \\ \hline\hline
\multirow{4}{*}{Input}      & $d$       & \begin{tabular}[c]{@{}l@{}}0.02 (m) for PointNet++,\\ 0.03 (m) for sparse CNN,\\ 0.04 (m) for KPConv\end{tabular} \\
                            & $R$       & 3 (m) \\
                            & $s$       & 3 (m) \\
                            & $k$       & 20000 \\
\multicolumn{1}{c}{} & $K$ & \begin{tabular}[c]{@{}l@{}}6 for PointNet++ and sparse CNN,\\ 7 for KPConv\end{tabular} \\ \hline
\multirow{3}{*}{Training}   & $W_{e}$   & 0.1 \\
                            & $W_{o}$   & 0.1 \\
                            & $W_{r}$   & 0 \\ \hline
\multirow{4}{*}{Thresholds} & $Th_d $   & $1.5\cdot{d}$ \\
                            & $Th_n$    & 32 \\
                            & $Th_{bm}$ & 0.01 \\
                            & $Bw$      & 0.6 \\ \hline
\end{tabular}
%}
\label{table:parameterSettingS3DIS}
\end{table}

%\begin{comment}
%\end{comment}

\begin{table}[thb]
\caption{Quantitative results for Area 5 of the S3DIS dataset. ``--'' means that the corresponding value is not available in the literature.}
\vspace{0.3cm}
\resizebox{\textwidth}{!}{
\begin{tabular}{ll|ccc|ccccc|ccc}
\hline
\multicolumn{2}{c|}{\multirow{2}{*}{method}}   & \multicolumn{3}{c|}{panoptic segmentation} & \multicolumn{5}{c|}{instance segmentation} & \multicolumn{3}{c}{semantic segmentation}                             \\ \cline{3-13} 
& & PQ & RQ & SQ & mCov & mWCov & mPrec & mRec & F1 & oAcc & mAcc & mIoU \\ \hline\hline
\multirow{2}{*}{PointNet++}
& offset      &22.0 &29.7 & 67.8& 45.3   & 46.1    & 26.2    & 45.2  & 33.2  & \multirow{2}{*}{86.0} & \multirow{2}{*}{65.9} & \multirow{2}{*}{58.7} \\ %\cline{3-7}
& embed       & 24.6 & 32.6 & 68.2 & 44.1 & 45.8 & 32.3 & 41.2  & 36.2  & & & \\ \cline{3-13} 
\multirow{2}{*}{Sparse CNN$\qquad$}
& offset      & 34.9 & 42.6 & 75.2  & 59.6 & 60.9 & 36.4 & 61.3  & 45.7  & \multirow{2}{*}{88.8} & \multirow{2}{*}{70.7} & \multirow{2}{*}{63.8} \\ %\cline{3-7}
& embed       &39.2 &48.0 & 74.9 & 56.2 & 57.9 & 49.9 & 56.2  & 52.9  & & & \\ \cline{3-13}
\multirow{2}{*}{KPConv}
& offset      & 33.7& 42.2& 73.3 & 58.6   & 59.6    & 37.4    & 59.5  & 46.0  & \multirow{2}{*}{89.2} & \multirow{2}{*}{71.3} & \multirow{2}{*}{65.3} \\ %\cline{3-7}
& embed       & 41.8 & 51.5 & 74.7 & 55.1 & 56.7 & 56.8 & 55.9  & 56.4  & & & \\ \hline
\multicolumn{2}{l|}{PointGroup (reprod.)} 
&42.3 & 52.0 & 74.7 & 56.8 & 57.7 & 55.3 & 56.6  & 56.0  & 89.2 & 71.2 & 64.9 \\
\multicolumn{2}{l|}{PointGroup (reprod., \sout{ScoreNet})}
& 35.6 & 44.5 & 74.0 & 56.6 & 57.5 & 41.8 & 55.8  & 47.8  & 89.0 & 70.4 & 64.2 \\ \hline \multicolumn{2}{l|}{3D-BoNet~\citep{Yang2019LearningOB}}
& -- & -- & -- & -- & -- & 57.5 & 40.2  & 47.3  & -- & -- & -- \\
\multicolumn{2}{l|}{ASIS~\citep{Wang2019AssociativelySI}}
& -- & -- & -- & 44.6 & 47.8 & 55.3 & 42.4  & 47.9  & 86.9 & 60.9 & 53.4 \\
\multicolumn{2}{l|}{MPNet~\citep{He2020LearningAM}}
& -- & -- & -- & 50.1 & 53.2 & 62.5 & 49.0  & 54.9  & -- & -- & -- \\
\multicolumn{2}{l|}{3D-MPA~\citep{Engelmann20CVPR}}
& -- & -- & -- & -- & -- & 63.1 & 58.0  & 60.4  & -- & -- & -- \\
\multicolumn{2}{l|}{PointGroup~\citep{jiang2020pointgroup}}
& -- & -- & -- & -- & -- & 61.9 & 62.1 & 61.9  & -- & -- & -- \\
\multicolumn{2}{l|}{ASNet~\citep{jiang2020end}}
& -- & -- & -- & 49.6 & -- & -- & -- & -- & -- & -- & -- \\ \hline
\end{tabular}
}
\label{table:S3DISArea5Compare}
\end{table}

\end{document}